%% file: main.tex
\documentclass[10pt,twocolumn,letterpaper]{article}

\usepackage[pagenumbers]{iccv} 

\definecolor{iccvblue}{rgb}{0.21,0.49,0.74}
\newcommand{\dname}[1]{TIP-I2V}
\newcommand{\dsname}[1]{TIPS-I2V}
\usepackage[pagebackref,breaklinks,colorlinks,allcolors=iccvblue]{hyperref}

\usepackage{amsmath}       
\usepackage{array}

\usepackage{xcolor}
\usepackage{tcolorbox}
\definecolor{lightroyalblue}{HTML}{F6F8FD}
\definecolor{mygray}{gray}{.9}
\definecolor{ggray}{RGB}{127,127,127}
\definecolor{reda}{RGB}{192,0,0}
\definecolor{redb}{RGB}{217,148,143}
\definecolor{myyellow}{RGB}{190,144,0}
\definecolor{mygreen}{RGB}{80,100,40}
\definecolor{myblue}{RGB}{30,90,100}
\definecolor{mygray1}{RGB}{245,245,245}
\usepackage{tabularx}
\newcolumntype{Y}{>{\centering\arraybackslash}X}
\usepackage{multirow}
\usepackage{caption}
\usepackage{graphicx}
\usepackage{tabularx}
\usepackage{colortbl}
\usepackage{fontawesome}
\usepackage{footmisc}

\definecolor{pp}{HTML}{964A6B}
\definecolor{bb}{HTML}{3476B9}
\makeatletter  
\newcommand{\thickhline}{%
	\noalign {\ifnum 0=`}\fi \hrule height 1pt
	\futurelet \reserved@a \@xhline
}
\newtcolorbox{abox}{colback=lightroyalblue,colframe=black,boxrule=0.5pt}
\usepackage{mdframed}
\def\paperID{40} 
\def\confName{ICCV}
\def\confYear{2025}

\title{\dname~: A Million-Scale Real Text and Image Prompt Dataset  \\ for Image-to-Video Generation\vspace{-0.5em}}
\author{%
  Wenhao Wang \\
  University of Technology Sydney\\
  {\tt\small wangwenhao0716@gmail.com} 
  \and
  Yi Yang\textsuperscript{*} \\
  Zhejiang University\\
  {\tt\small yangyics@zju.edu.cn}
}

\begin{document}
\twocolumn[{%
\renewcommand\twocolumn[1][]{#1}%
\maketitle
\begin{center}
    \centering
    \vspace{-1.9em}
    \includegraphics[width=\textwidth]{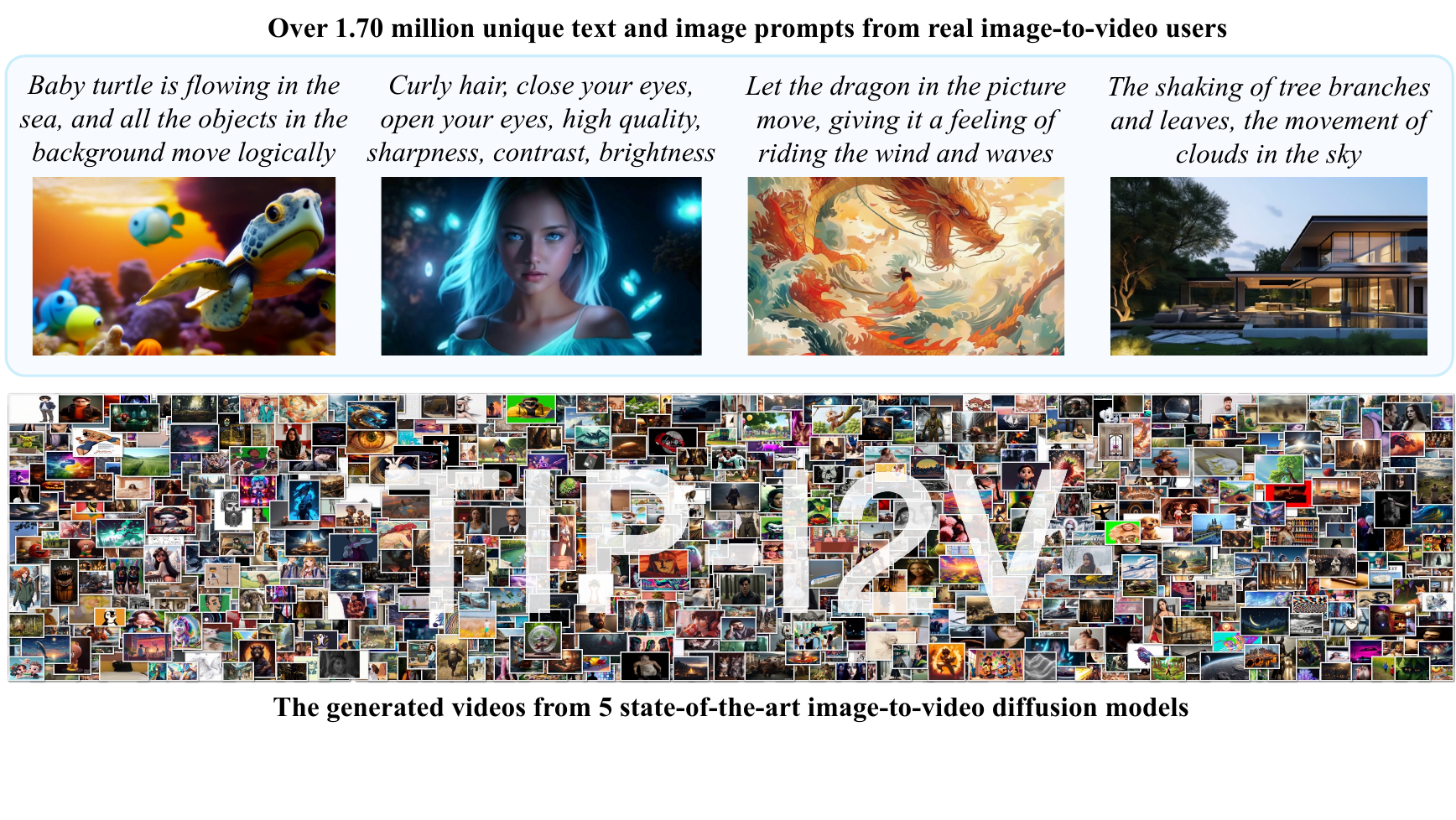}
    \vspace{-4.7em}
    \captionof{figure}{\textbf{\dname~} is the first dataset comprising over 1.70 million unique user-provided text and image prompts. Besides the prompts, \dname~ also includes videos generated by five state-of-the-art image-to-video models ($\mathtt{Pika}$ \cite{pika2024}, $\mathtt{Stable}$ $\mathtt{Video}$ $\mathtt{Diffusion}$ \cite{blattmann2023stable}, $\mathtt{Open\text{-}Sora}$ \cite{opensora}, $\mathtt{I2VGen\text{-}XL}$ \cite{zhang2023i2vgen}, and $\mathtt{CogVideoX\text{-}5B}$ \cite{yang2024cogvideox}). The \dname~ inspires new research directions related to image-to-video generation.}
    \label{Fig: teasor}
\end{center}
}]

\setcounter{footnote}{0}
\footnotetext{\textsuperscript{*}Corresponding author.}

\begin{abstract}
\vspace{-5mm}

Video generation models are revolutionizing content creation, with image-to-video models drawing increasing attention due to their enhanced controllability, visual consistency, and practical applications. However, despite their popularity, these models rely on user-provided text and image prompts, and there is currently no dedicated dataset for studying these prompts.
In this paper, we introduce \textbf{\dname~}, the first large-scale dataset of over $1.70$ million unique user-provided \textbf{T}ext and \textbf{I}mage \textbf{P}rompts specifically for \textbf{I}mage-to-\textbf{V}ideo generation. Additionally, we provide the corresponding generated videos from five state-of-the-art image-to-video models. We begin by outlining the time-consuming and costly process of curating this large-scale dataset. Next, we compare \dname~ to two popular prompt datasets, VidProM (text-to-video) and DiffusionDB (text-to-image), highlighting differences in both basic and semantic information. This dataset enables advancements in image-to-video research. For instance, to develop better models, researchers can use the prompts in \dname~ to analyze user preferences and evaluate the multi-dimensional performance of trained models; and to enhance model safety, they may focus on addressing the misinformation issue caused by image-to-video models. The new research inspired by \dname~ and the differences with existing datasets emphasize the importance of a specialized image-to-video prompt dataset.
The project is available at \textcolor{iccvblue}{https://tip-i2v.github.io/}.

\end{abstract}

\vspace{-4mm}
\section{Introduction}
\hspace{1.2em}Image-to-video diffusion models transform static images into dynamic videos, with wide-ranging applications in animation, content creation, and visual storytelling \cite{ni2023conditional,xing2023survey,guo2024i2v,zhang2024trip,zhang2023i2vgen}. As video generation becomes more commercialized, \textbf{image-to-video} methods are increasingly preferred over \textcolor[RGB]{100, 100, 100}{text-to-video} for several reasons: (1) they offer users more \textbf{control}, allowing for precise direction of objects to perform specific actions; (2) they provide greater \textbf{consistency}, enabling narratives focused on a central subject; and (3) they are more \textbf{practical}, particularly on social media where image-driven videos tend to gain higher engagement.
\par 

Despite their popularity and importance, there currently lacks a dataset from the user’s perspective -- one that features user-provided text and image prompts alongside the corresponding generated videos. Such a dataset could help improve the alignment of image-to-video models with real-world user needs, while also enhancing safety.
Therefore, this paper conducts the first study of its kind in the image-to-video community. Specifically, we focus on curating the \textit{first} image-to-video prompt-gallery dataset, analyzing the \textit{differences} between the proposed dataset and similar ones, and exploring the \textit{new research directions} inspired by us. \par

\textbf{The first Text and Image Prompt dataset for Image-to-Video generation (\dname~).} As shown in Fig. \ref{Fig: teasor}, our \dname~ dataset includes over $1.70$ million unique user-provided text and image prompts for image-to-video diffusion models, along with the corresponding generated videos, sourced from Pika Discord channels \cite{pika2024}. It is important to note that: 
\textbf{(1)} We intend to include videos generated by other state-of-the-art image-to-video diffusion models, including $\mathtt{Stable}$ $\mathtt{Video}$ $\mathtt{Diffusion}$ \cite{blattmann2023stable}, $\mathtt{Open\text{-}Sora}$ \cite{opensora}, $\mathtt{I2VGen\text{-}XL}$ \cite{zhang2023i2vgen}, and $\mathtt{CogVideoX\text{-}5B}$ \cite{yang2024cogvideox}; however, due to limited computing resources, we only use $100,000$ randomly selected prompts to generate videos for each image-to-video model.
Researchers are free to extend our \dname~ by generating more videos with these methods (or other state-of-the-arts, such as $\mathtt{Open\text{-}Sora\text{-}Plan}$ \cite{pku_yuan_lab}) and our prompts; 
\textbf{(2)} we acknowledge that the currently generated videos are not perfect, and in the future, researchers are encouraged to use newly released image-to-video models (such as $\mathtt{Sora}$ \cite{openai2024sora}) and our prompts to further extend our \dname~. Besides prompts and generated videos, our \dname~ also includes Universally Unique Identifier (UUIDs), anonymous UserIDs, timestamps, embeddings, subjects, and not-safe-for-work (NSFW) scores for these data points. 

\textbf{Differences between \dname~ and other similar datasets in basic and semantic information.} We notice that there are two popular prompt-gallery datasets in the visual generation community, \ie, VidProM \cite{wang2024vidprom} for \textit{text-to-video} and DiffusionDB \cite{wang2023diffusiondb} for \textit{text-to-image}. Our \dname~ mainly differs from them in: \textbf{(1) Basic information:} Both VidProM and DiffusionDB begin the generation with a \textit{text}, while our \dname~ starts with a \textit{text} and an \textit{image}. \textbf{(2) Semantics:} Each text in our \dname~ focuses on how to bring static elements in the corresponding image to life through motion. For instance, some text prompts are ``\textit{the hair and body dance}", ``\textit{the flowers fall, the woman move}", and ``\textit{make the statue break down}". In contrast, the prompts in VidProM and DiffusionDB are more \textit{descriptive}, \ie, they directly specify the content to be generated without referencing a specific object, such as ``\textit{a dragon flies over a city}'', ``\textit{a beach at sunset}'', and ``\textit{an astronaut walks on Mars}''.
The \textbf{differences} in basic and semantic information 
highlight a need for an image-to-video prompt dataset.\par 

\textbf{Exciting new research areas inspired by \dname~.}
Our \dname~ helps researchers develop \textit{better} and \textit{safer} image-to-video diffusion models. For \textbf{better} models: \textbf{(1) Enhancing user experience.} Before \dname~, researchers do not know which \textit{subjects} users prefer to transform into videos and what \textit{directions} they expect. However, after analyzing our \underline{prompts}, researchers can collect videos from YouTube according to users' preferred subjects and directions. This avoids the previous problem, \ie, blindly expanding the training set led to wasted resources and low user satisfaction. \textbf{(2) Improving evaluation practicality.} Existing image-to-video benchmarks, such as $\mathtt{VBench}$-$\mathtt{I2V}$ \cite{huang2023vbench}, $\mathtt{I2V}$-$\mathtt{Bench}$ \cite{renconsisti2v}, and $\mathtt{AIGCBench}$ \cite{fan2024aigcbench}, suffer from a limited number of topics and prompts designed by experts, which may not accurately reflect real-world user needs. With the help of \underline{prompts} in \dname~, researchers can build a more \textit{comprehensive} and \textit{practical} benchmark for evaluating image-to-video models.
For \textbf{safer} models: A major safety concern of image-to-video generation is \textbf{misinformation}, \ie, they can make objects or humans in images to perform actions they never did. For instance, given an image of $\mathtt{Elon}$ $\mathtt{Musk}$ and $\mathtt{Donald}$ $\mathtt{Trump}$ together, image-to-video models could generate a video of them fighting, misleading the public. To address this issue, \underline{videos} in our \dname~ allows researchers to train a model to \textbf{(1)} distinguish between generated videos from images and real videos, and \textbf{(2)} trace the source image from any given frame in a generated video.
Beyond these areas, we also encourage researchers to explore \textbf{additional directions}.\par
\begin{figure*}[t]
    \centering
    \includegraphics[width=\textwidth]{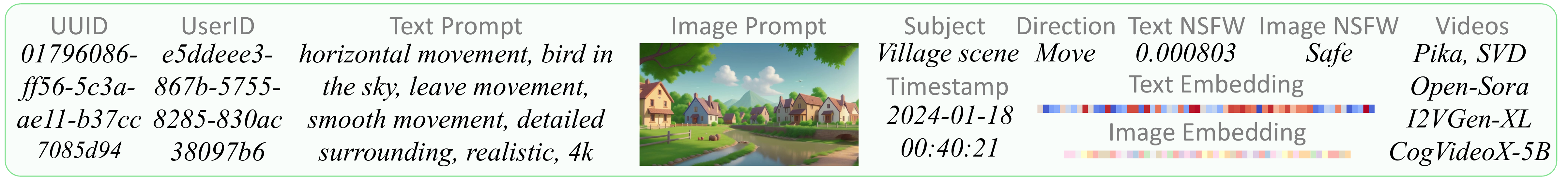}
    \vspace{-6mm}  
    \caption{A data point in our \textbf{\dname~} includes UUID, anonymous UserID, text and image prompt, timestamp, subject and direction, NSFW status of text and image, text and image embedding, and the corresponding generated videos.} 
    \label{Fig: data}
   \vspace{-4mm} 
\end{figure*}

In conclusion, our key contributions are as follows:

\begin{enumerate}

\item We present \dname~, the first dataset of text and image prompts specifically for image-to-video generation. This dataset contains over $1.70$ million prompts from real users, along with corresponding videos generated by five state-of-the-art image-to-video diffusion models.

\item We compare \dname~ with two popular prompt datasets, \ie, VidProM (text-to-video) and DiffusionDB (text-to-image), highlighting their differences in both basic and semantic information. This emphasizes the need for a specialized image-to-video prompt dataset.

\item We demonstrate how \dname~ contributes to building better and safer image-to-video diffusion models. Specifically, it can help enhance user experience, improve the practicality of evaluation, distinguish generated videos from real ones, and trace the source image.

\end{enumerate}

\section{Related Works}

\hspace{1.2em}\textbf{Video Generation.}
With the introduction of $\mathtt{Sora}$ \cite{openai2024sora}, there has been increasing interest in video generation, leading to many real-world applications. Previous researchers \cite{text2video-zero,chen2023videocrafter1,chen2024videocrafter2,wang2023modelscope} have primarily focused on \textit{text-to-video} generation, exploring methods for synthesizing realistic videos directly from textual descriptions. However, a main drawback of text-to-video is \textit{uncontrollability} and \textit{inconsistency}. For instance, the short film \textit{Air Head} produced by $\mathtt{Sora}$ \cite{openai2024sora} takes professional film crews a long time to control that the man has a consistent yellow balloon head. A promising solution is \textit{image-to-video} generation, where a video is created from an image, and text is used to control the objects within the image. This makes the generated videos more consistent.
The videos created from images by recent works such as 
\cite{blattmann2023stable,zhang2023i2vgen,kuaishou_kling,hailuo_ai_video} are rapidly gaining popularity on social media platforms.
This paper also focuses on image-to-video generation, but from a different perspective, \ie, researching user-provided text and image prompts.

\textbf{Text-Video Datasets.}
A text-video dataset is a collection of video clips with corresponding textual descriptions or captions. There are many popular text-video datasets, such as $\mathtt{HDVILA}$-$\mathtt{100M}$ \cite{xue2022advancing}, $\mathtt{WebVid}$-$\mathtt{10M}$ \cite{bain2021frozen}, $\mathtt{Panda}$-$\mathtt{70M}$ \cite{chen2024panda}, $\mathtt{InternVid}$ \cite{wanginternvid}, $\mathtt{OpenVid}$-$\mathtt{1M}$ \cite{nan2024openvid}, and $\mathtt{MiraData}$ \cite{ju2024miradata}, offering high-resolution and diverse content. Unlike other datasets that consist of caption-(real)-video pairs, our \dname~ dataset contains real user-provided text and image prompts, along with the corresponding generated videos.
To further highlight the difference, in the Supplementary (Section \ref{sec:tvdataset}), we also provide a \textsc{WizMap} visualization comparing the texts in our \dname~ with those in $\mathtt{Panda}$-$\mathtt{70M}$ \cite{chen2024panda}.

\textbf{Prompt Datasets.} With the rapid spread of large AI models (such as large language models and diffusion models), research on \textit{prompts} has become particularly important, as they form the foundation of efficient user interactions with AI systems. Based on this background, in the \textit{text-to-text} community, \cite{zhong2021adapting} aggregates $43$ existing datasets from various domains and tasks to create a prompt dataset, while \cite{bach2022promptsource} develops PromptSource, a system for creating, sharing, and managing prompts for natural language processing (NLP) tasks. In the visual generation domain, VidProM \cite{wang2024vidprom} and DiffusionDB \cite{wang2023diffusiondb} collect prompts for \textit{text-to-video} and \textit{text-to-image} tasks, respectively. However, to the best of our knowledge, there are \textbf{no} prompt datasets for the image-to-video task. Given the importance and popularity of this task, our paper aims to fill this gap.

\begin{table*}[t]
\centering
\begin{minipage}{1.0\textwidth}
\captionsetup{font=small}
\caption{Comparison of our \textbf{\dname~} (image-to-video) with popular VidProM (text-to-video) and DiffusionDB (text-to-image) in terms of \textbf{basic} information. Our \dname~ is comparable in scale to these datasets but focuses on different aspects of visual generation.} 
\vspace*{-2mm}
\hspace*{-2mm}
\small
\scalebox{1}{
  \begin{tabularx}{\hsize}{|>{\raggedleft\arraybackslash}p{4.2cm}||>{\centering\arraybackslash}p{4.2cm}|>{\centering\arraybackslash}p{4.2cm}|Y|}
    \hline\thickhline
    \rowcolor{mygray}\multirow{1}{*}{Details}
 &\multicolumn{1}{c|}{\textbf{\dname~ (Ours)}} 
     & \multicolumn{1}{c|}{VidProM \cite{wang2024vidprom}}& \multicolumn{1}{c|}{DiffusionDB \cite{wang2023diffusiondb}} \\ 
    
    \hline\hline
    Domain &\cellcolor{lightroyalblue}Image-to-Video&Text-to-Video&Text-to-Image\\
     No. of unique \textit{text} prompts &\cellcolor{lightroyalblue}$1,701,935$& $1, 672, 243$ & $1, 819, 808$\\ 
      No. of unique \textit{image} prompts &\cellcolor{lightroyalblue}$1,701,935$& - & - \\
      Embedding of \textit{text} prompts &\cellcolor{lightroyalblue}$\mathtt{text}$-$\mathtt{embedding}$-$\mathtt{3}$-$\mathtt{large}$& $\mathtt{text}$-$\mathtt{embedding}$-$\mathtt{3}$-$\mathtt{large}$ & $\mathtt{CLIP}$ \\
      Embedding of \textit{image} prompts &\cellcolor{lightroyalblue}$\mathtt{CLIP}$& - & - \\
      Maximum length of  \textit{text} prompts&\cellcolor{lightroyalblue}$8192$ tokens& $8192$ tokens & $77$ tokens \\
      Time span&\cellcolor{lightroyalblue}Jul 2023 $\sim$ Oct 2024& Jul 2023 $\sim$ Feb 2024 & Aug 2022 \\
      No. of generation sources &\cellcolor{lightroyalblue}$5$&$4$&$1$ \\
      Collection method&\cellcolor{lightroyalblue}Web scraping + Local generation& Web scraping + Local generation & Web scraping \\
    \hline
  \end{tabularx}}
  \label{Table: Dataset}
  \vspace*{-4mm}
  \end{minipage}
\end{table*}

\section{Curating \dname~}
\hspace{1.2em}Fig. \ref{Fig: data} is a data point in the \dname~. It includes a UUID, an anonymous UserID, text and image prompts, timestamp, subject and direction, NSFW status for text and images, embeddings for both text and images, and generated videos from five state-of-the-art image-to-video models. The following explains how we collect these pieces of information.

\textbf{Collecting source HTML files.} We collect chat messages from Pika’s official Discord channels between July 2023 and October 2024 by using DiscordChatExporter \cite{DiscordChatExporter} to save them as HTML files. According to Pika's terms of service (see exact words in the Supplementary (Section \ref{Sup: License})), these chat messages are publicly available under the CC BY-NC 4.0 License. As a result, we comply with this license and release our \dname~ under the same terms. This collection process is similar to that of the well-established datasets VidProM \cite{wang2024vidprom} and DiffusionDB \cite{wang2023diffusiondb}.

\textbf{Extracting text prompts, scraping Pika videos, and parsing image prompts.} We use regular expressions to extract text prompts and their corresponding video links from HTML files. After deduplicating the text prompts and validating the video links, we obtained $1,701,935$ unique text prompts along with their corresponding $3$s-length Pika videos. Since the original image prompts are not accessible, and the image-to-video models developed by Pika utilize these user-inputted prompts as the first frames of generated videos, the image prompts in our \dname~ are parsed from the scraped videos. We have checked the quality of these image prompts and find that they are of high quality.

\textbf{Assigning UUIDs, UserIDs, timestamps, subjects, embeddings and NSFW status.} To facilitate subsequent research utilizing our \dname~, we (1) calculate UUIDs and anonymous UserIDs based on unique prompts and user names, respectively, (2) extract timestamps from Pika videos, (3) infer subjects and directions using $\mathtt{GPT}$-$\mathtt{4o}$ \cite{openai2024hello}, (4) embed text and image prompts using $\mathtt{text}$-$\mathtt{embedding}$-$\mathtt{3}$-$\mathtt{large}$ \cite{openai2024embedding} and $\mathtt{CLIP}$ \cite{radford2021learning}, respectively, and (5) assign NSFW status to text and image prompts using $\mathtt{Detoxify}$ \cite{Detoxify} and $\mathtt{nsfw\_image\_detection}$ \cite{huggingface2024nsfw}, respectively.


\textbf{Generating videos using other image-to-video models.} To diversify the proposed \dname~, we also include videos generated by other state-of-the-art diffusion models, \ie, $\mathtt{Stable}$ $\mathtt{Video}$ $\mathtt{Diffusion}$ \cite{blattmann2023stable}, $\mathtt{Open\text{-}Sora}$ \cite{opensora}, $\mathtt{I2VGen\text{-}XL}$ \cite{zhang2023i2vgen}, and $\mathtt{CogVideoX\text{-}5B}$ \cite{yang2024cogvideox}. See the Supplementary (Section \ref{Supple: i2v}) for details on how we adopt these models. 
Due to the time constraint (for instance, generating a single video with $\mathtt{CogVideoX\text{-}5B}$ \cite{yang2024cogvideox} requires $294$ seconds on a standard A100 GPU), we limit the number of generated videos for each diffusion model to $100,000$. This number of generated videos is likely to be sufficient for drawing conclusions in subsequent research.

\textbf{Extension.} The large-scale collection of user-provided prompts enables future researchers to extend the \dname~. \textbf{F}irst, if they currently need more generated videos, they can generate additional videos using the diffusion models we employed or other state-of-the-art models. \textbf{A}dditionally, as more advanced image-to-video models become available in the future, researchers can leverage these models with our prompts to generate higher-quality videos.

\section{Comparing \dname~ with Similar Datasets}
\begin{figure*}[t]
    \centering
    \includegraphics[width=0.99\textwidth]{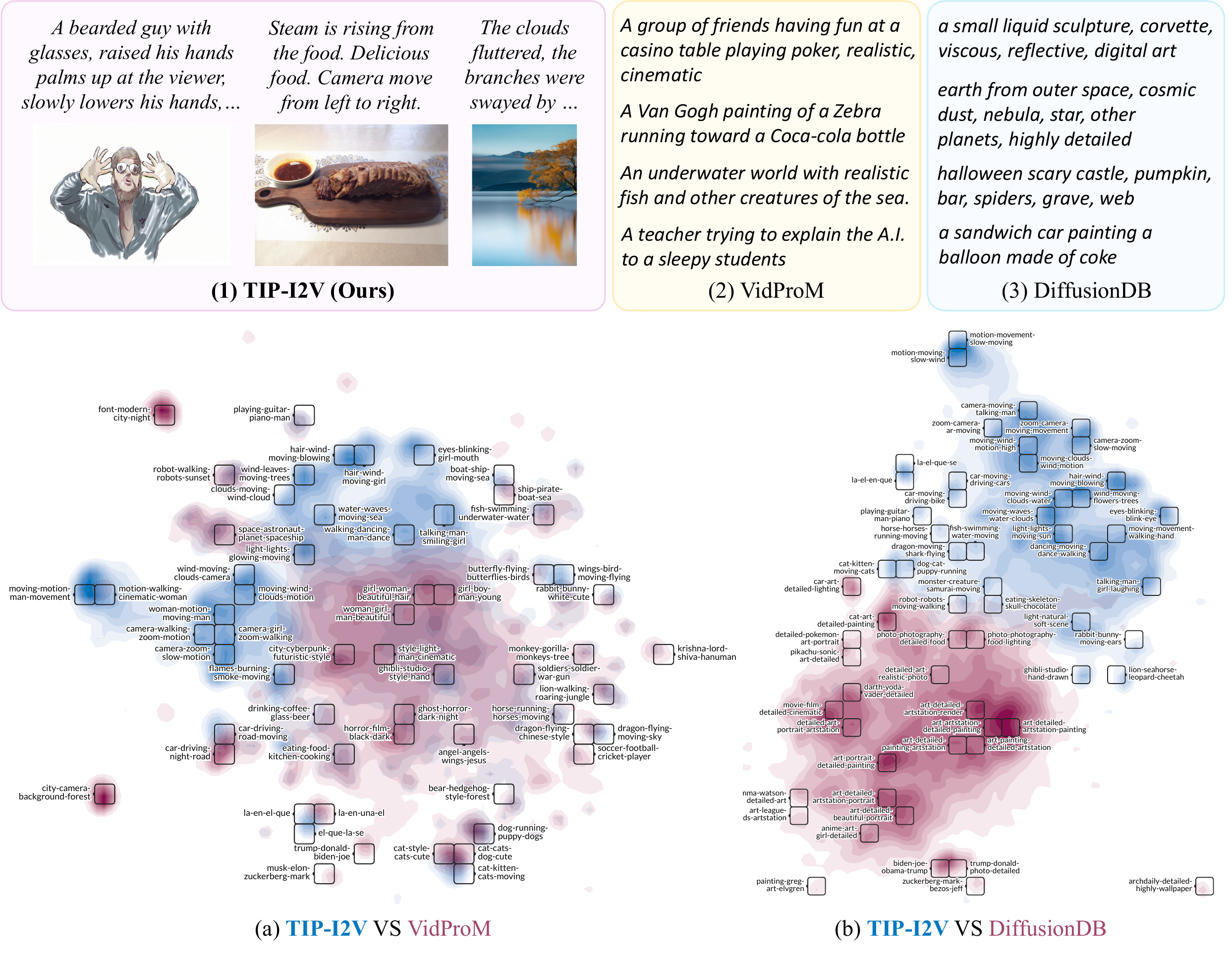}
    \vspace{-5mm}  
    \caption{Our \textbf{\dname~} (\textit{image-to-video}) differs from popular VidProM \cite{wang2024vidprom} (\textit{text-to-video}) and DiffusionDB \cite{wang2023diffusiondb} (\textit{text-to-image}) in terms of \textbf{semantics}. \textbf{Top:} Example prompts from the three datasets. \textbf{Bottom:} The \textsc{WizMap} \cite{wang2023wizmap} visualization of our \textbf{\textcolor{bb}{\dname~}} compared to \textcolor{pp}{VidProM/DiffusionDB}. Please \faSearch~zoom in to see the detailed semantic focus of text prompts across the three datasets.} 
    \label{Fig: comparison}
   \vspace{-4mm} 
\end{figure*}

\hspace{1.2em}In this section, we compare the proposed \dname~ with similar datasets, \ie, VidProM \cite{wang2024vidprom} and DiffusionDB \cite{wang2023diffusiondb}, in terms of \textit{basic information} and \textit{prompt semantics}. The differences highlight the necessity of introducing a specialized prompt dataset for image-to-video generation.

\textbf{Basic information.} As shown in Table \ref{Table: Dataset}, we compare the basic information of \dname~ with two existing prompt-gallery datasets, \ie, VidProM \cite{wang2024vidprom} (text-to-video) and DiffusionDB \cite{wang2023diffusiondb} (text-to-image). It is observed that: \textbf{(1)} All three datasets have reached a million-scale,  providing support for the research of large-scale machine learning. \textbf{(2)} Unlike these datasets, which only contain text prompts, our \dname~ also includes image prompts. \textbf{(3)} Our dataset spans a longer collection period and includes more generation sources compared to these datasets, which suggests a greater diversity of prompts and generated outputs.
\vspace{-1mm}
\begin{abox} 
\vspace{-1mm} 
    \looseness -1 \textbf{Takeaway:} The \dname~ and popular datasets are all large-scale, but we focus on a different domain, which additionally needs image prompts.
\vspace{-1mm} 
\end{abox}
\vspace{-1mm}
\textbf{Prompt semantics.}
As shown in Fig. \ref{Fig: comparison}, we present a comparison of the prompts used in our \dname~, VidProM \cite{wang2024vidprom}, and DiffusionDB \cite{wang2023diffusiondb}. Additionally, we compare the semantic distributions of the prompts across these datasets. From the comparisons, we conclude that: \textbf{(1)} The text prompts in our \dname~ serve as instructions to animate the specified object(s) described in the corresponding image prompts. For instance, on the left side of Fig. \ref{Fig: comparison} \textcolor{iccvblue}{(1)}, the user wants to see ``\textit{the bearded guy}" ``\textit{lower his hands}". In contrast, the prompts in VidProM and DiffusionDB primarily describe the scenes that users expect to visualize. The drawback is that no matter how detailed users specify, the generation may always has visual differences from the images/videos they have in mind. \textbf{(2)} The \textsc{WizMap} \cite{wang2023wizmap} visualization shows that the text prompts in \dname~ have a distinct distribution compared to those in VidProM and DiffusionDB. This further validates the semantic differences between our text prompts and those in existing datasets. 

\vspace{-1mm}
\begin{abox} 
\vspace{-1mm} 
    \looseness -1 \textbf{Takeaway:} The semantics of the text prompts in our \dname~ differs from those in popular datasets.
 \vspace{-1mm} 
\end{abox}
\vspace{-1mm}

\section{New Research based on \dname~}

\hspace{1.2em}In this section, we first elaborate on four new research directions inspired by our \dname~, followed by a brief discussion of other potential ones. These directions collaboratively contribute to improving the \textit{quality} and \textit{safety} of image-to-video generation.

\subsection{Catering Users Better}
\begin{figure}[t]

    \centering
    \hspace{-2mm}
    \includegraphics[width=0.48\textwidth]{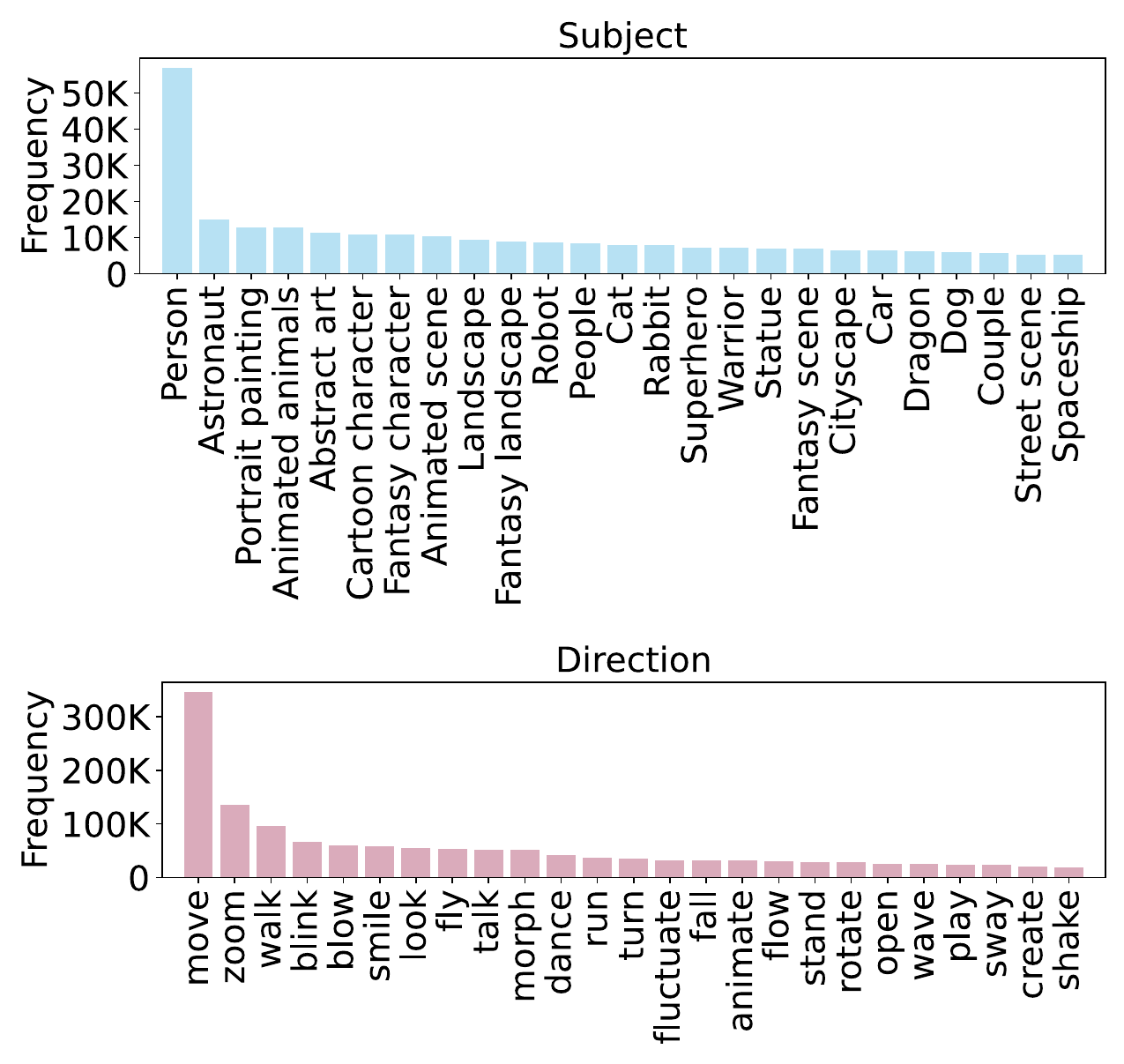}
    \vspace{-8mm}  
    \caption{The top $25$ \textit{subjects} (top) and \textit{directions} (bottom) preferred by users when generating videos from images.} 
    \label{Fig: sort}
   \vspace{-2mm} 
\end{figure}

\begin{figure}[t]
    \centering
    \includegraphics[width=0.48\textwidth]{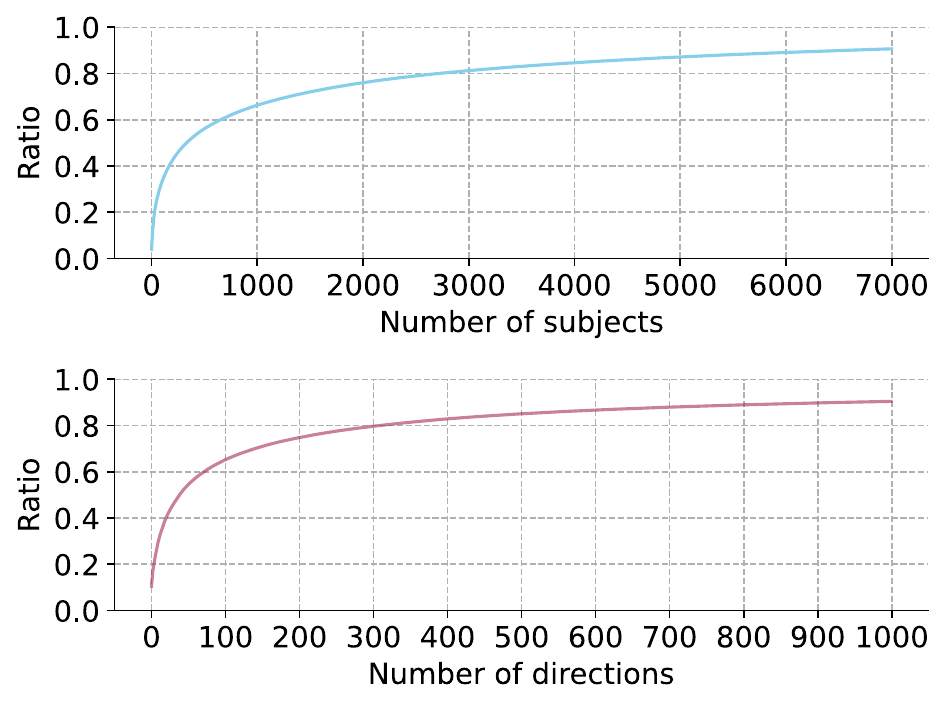}
    \vspace{-8mm}  
    \caption{The ratio of the sum of the top $N$ \textit{subjects} (top) or \textit{directions} (bottom)  to the total frequencies.} 
    \label{Fig: ratio}
   \vspace{-4mm} 
\end{figure}

\hspace{1.2em}\textbf{Analysis of users’ preferred subjects and generation directions.} In Fig. \ref{Fig: sort}, we visualize the top $25$ subjects and directions preferred by users. Calculation details and examples are provided in the Supplementary (Sections \ref{Supple: User} and \ref{Supple: Example}). Furthermore, in Fig. \ref{Fig: ratio}, we show the proportion of the sum of top $N$ subjects/directions relative to the total frequencies. We observe that: \textbf{(1)} The users' preferences are \textbf{unbalanced}. For instance, the top-3 subjects, \ie, \textit{``person"}, \textit{``astronaut"}, and \textit{``portrait painting"}, are all human-related. Beyond the general action \textit{``move''}, people are more likely to generate specific movements such as \textit{``zoom''}, \textit{``walk''}, and \textit{``blink''}. \textbf{(2)} To cater to the preferences of the general audience, image-to-video model designers only need to focus on \textbf{a relative small number} of subjects and directions. Specifically, to cover $80\%$/$90\%$ users' preferences,
researchers only need to focus on $2,721$/$6,586$ subjects and $309$/$929$ directions, respectively. 

These two observations imply that, to create a successful commercial image-to-video model, researchers may only need to focus on these specific subjects and directions, rather than wasting resources and time to expand datasets blindly. 
Therefore, future research may focus on:

\textbf{$\bullet$ User-oriented training datasets for video generation.} In the future, researchers may search for top subjects on free-to-use video platforms and scrape the resulting videos. Unlike previous large-scale datasets, which were randomly collected from websites, datasets curated through this procedure will better align with users' expectations, and models trained on them are likely to gain more popularity.

\textbf{$\bullet$ More precise semantic segmentation for subjects.} Although the subjects in our \dname~ are inferred by the powerful $\mathtt{GPT}$-$\mathtt{4o}$ \cite{openai2024hello}, some semantic overlap is inevitable. For instance, both \textit{``person''}, \textit{``people''}, and \textit{``man''} appear as subjects. This may negatively impact the construction of user-oriented video datasets, and therefore researchers may design methods to minimize semantic overlap within subjects before scraping.

\subsection{More Comprehensive and Practical Evaluation}

\hspace{1.2em}\textbf{Based on \dname~, we can conduct a more comprehensive and practical evaluation of image-to-video models.} As shown in Table \ref{Table: Bench}, the current benchmarks for image-to-video generation face two main issues. \textbf{Issue 1: comprehensiveness.} Their benchmarks cover only a limited range of subjects, which may result in the omission of many topics of interest. \textbf{Issue 2: practicality.} Their image prompts are directly extracted from frames in the public videos, and their text prompts are generated by multimodal models. This may not accurately reflect the needs of real-world users and differs from their actual usage habits.\par 

\begin{figure}[t]
    \centering
    \hspace*{-2mm}  
    \includegraphics[width=0.49\textwidth]{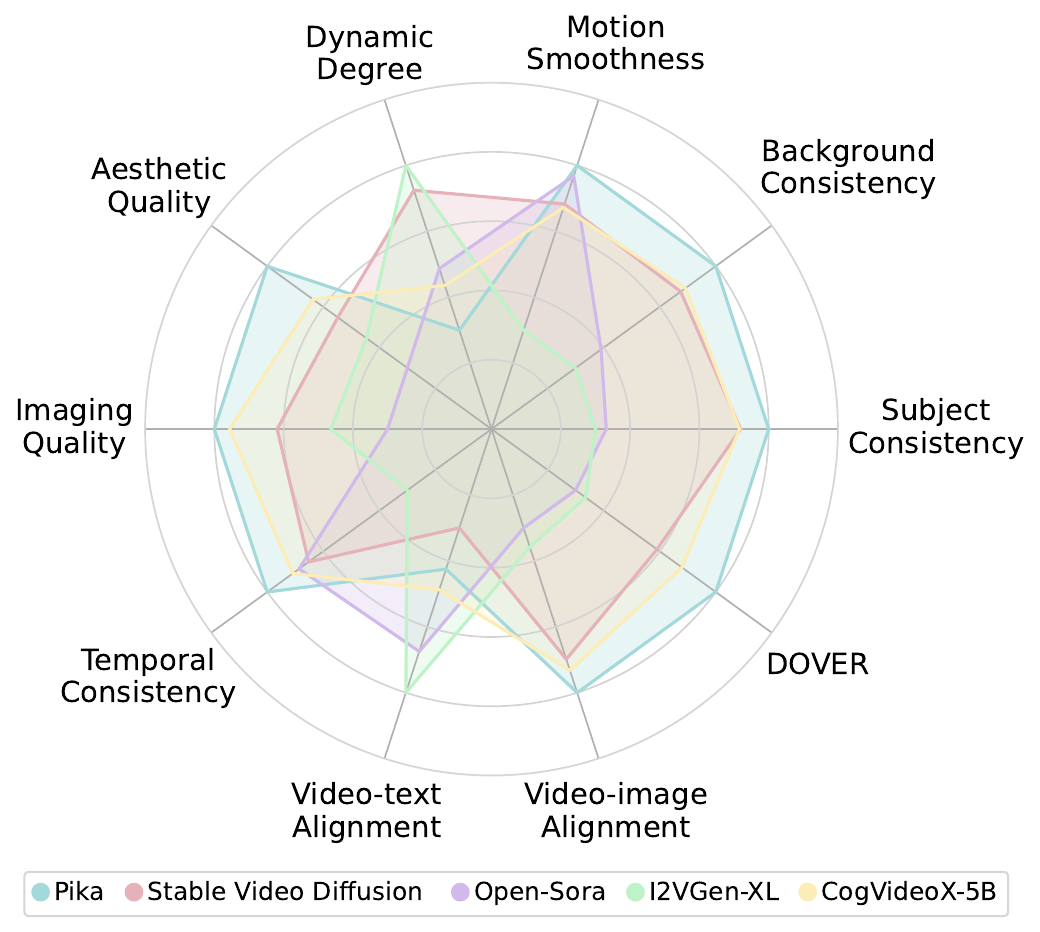}
    \vspace{-7mm}  
    \caption{Benchmarking results using $10,000$ prompts in TIP-Eval and 10 dimensions from \cite{huang2023vbench, renconsisti2v, fan2024aigcbench}. Similar to VBench \cite{huang2023vbench}, results are normalized per dimension for clearer comparisons.} 
    \label{Fig: bench}
   \vspace{-5mm} 
\end{figure}

To solve these problems, we propose \textbf{TIP-Eval}, a benchmark consisting of $1,000$ of the most popular subjects, each paired with $10$ text and image prompts provided by real users. Using these comprehensive and practical prompts along with the evaluation dimensions from \cite{huang2023vbench, renconsisti2v, fan2024aigcbench}, we benchmark the videos generated by five state-of-the-art image-to-video diffusion models. The visualization results are shown in Fig. \ref{Fig: bench}; for the full experiments, please refer to the Supplementary (Section \ref{Supple: Bench}).
We observe that: \textbf{(1)} From the user’s perspective, even the early-stage commercial image-to-video model ($\mathtt{Pika}$ \cite{pika2024}) outperforms the latest open-source one ($\mathtt{CogVideoX\text{-}5B}$ \cite{yang2024cogvideox}) on the majority of dimensions ($8$ out of $10$). This may somewhat contrast with the evaluation based on expert-designed prompts (as shown in the $\mathtt{CogVideoX\text{-}5B}$ paper), which emphasizes the practicality and importance of our TIP-Eval. \textbf{(2)} No image-to-video model outperforms across all dimensions, indicating the complexity of balancing different evaluation dimensions, such as \textit{consistency}, \textit{dynamic}, and \textit{alignment}. \textbf{(3)} The performance of all models on the \textit{video-text alignment} dimension is suboptimal, with the highest score being only $0.26$. 
This indicates that current image-to-video models still struggle to accurately adhere to human control.

Beyond the current analysis, in the future, researchers can use our TIP-Eval and \dname~ to explore:\par 
\begin{table}[t]
\centering
\begin{minipage}{0.485\textwidth}
\captionsetup{font=small}
\caption{A comparison of the proposed benchmark with existing ones. Our TIP-Eval is more \textbf{comprehensive} and \textbf{practical}.} 
\vspace*{-2mm}
\hspace*{-1.5mm}
\small
  \begin{tabularx}{\hsize}{|>{\raggedleft\arraybackslash}p{2.3cm}||>{\centering\arraybackslash}p{1.1cm}|>{\centering\arraybackslash}p{1.1cm}|>{\centering\arraybackslash}p{1.1cm}|Y|}
    \hline\thickhline
    \rowcolor{mygray}
 &\multicolumn{2}{c|}{Number} 
    & \multicolumn{2}{c|}{Prompt source}\\ 
   \rowcolor{mygray}\multirow{-2}{*}{Benchmarks}&\multicolumn{1}{c|}{Subjects} 
     & \multicolumn{1}{c|}{Prompts}& \multicolumn{1}{c|}{Image}& \multicolumn{1}{c|}{Text} \\ 
    \hline\hline
    VBench-I2V \cite{huang2023vbench} &$11$&$355$&Pexels&Gen.\\
    I2V-Bench \cite{renconsisti2v} &$16$&$2,951$&YouTube&Gen.\\
    AIGCBench \cite{fan2024aigcbench} &-&$1,000$&WebVid&Gen.\\
    \cellcolor{lightroyalblue} \textbf{TIP-Eval  (Ours)} & \cellcolor{lightroyalblue}$\mathbf{1,000}$& \cellcolor{lightroyalblue}$\mathbf{10,000}$& \multicolumn{2}{c|}{\cellcolor{lightroyalblue}\textbf{Real users}}\\
    \hline
  \end{tabularx}
  \label{Table: Bench}
  \vspace*{-3mm}
  \end{minipage}
\end{table}

\begin{figure*}[t]
    \centering
    \hspace*{-2mm}  
    \includegraphics[width=1.0\textwidth]{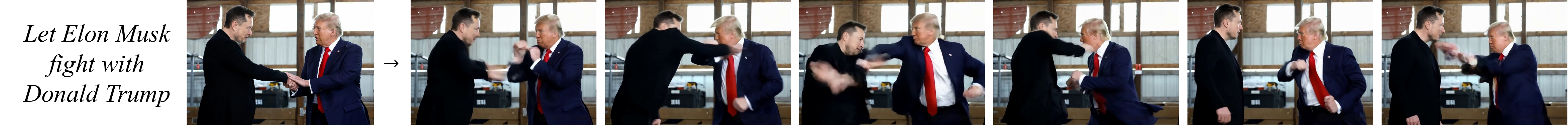}
    \vspace{-2mm}  
    \caption{A case illustrating the \textbf{misuse} of image-to-video models, resulting in \textbf{misinformation}: given a friendly image of $\mathtt{Elon}$ $\mathtt{Musk}$ and $\mathtt{Donald}$ $\mathtt{Trump}$ shaking hands, an image-to-video model can easily generate a video of them fighting, which fuels \textbf{political rumors}.} 
    \label{Fig: fight}
   \vspace{-5mm} 
\end{figure*}

\begin{table}[t]
\centering
\begin{minipage}{0.485\textwidth}
\captionsetup{font=small}
\caption{The \textbf{generalization} experiments of existing fake image detection methods to identify generated videos from images.} 
\vspace*{-2mm}
\hspace*{-1.5mm}
\small
  \begin{tabularx}{\hsize}{|>{\raggedleft\arraybackslash}p{1.7cm}||Y|Y|Y|Y|Y|Y|}
    \hline\thickhline
   \rowcolor{mygray}\multirow{1}{*}{\scalebox{0.95}{Accuracy ($\%$)}}&\multicolumn{1}{c|}{\scalebox{0.95}{Pika}} 
     & \multicolumn{1}{c|}{\scalebox{0.95}{SVD}}& \multicolumn{1}{c|}{\scalebox{0.95}{OpS}}& \multicolumn{1}{c|}{\scalebox{0.95}{IXL}} &\multicolumn{1}{c|}{\scalebox{0.95}{Cog}} &\multicolumn{1}{c|}{\scalebox{0.95}{Avg.}}\\ 
    \hline\hline
    \scalebox{0.95}{Blind Guess} &\scalebox{0.95}{$50.0$}&\scalebox{0.95}{$50.0$}& \scalebox{0.95}{$50.0$}&\scalebox{0.95}{$50.0$}&\scalebox{0.95}{$50.0$}&\cellcolor{lightroyalblue}\scalebox{0.95}{$50.0$}\\\hline\hline
    \scalebox{0.95}{\scalebox{0.95}{CNNSpot} \cite{wang2020cnn}} &\scalebox{0.95}{$50.7$}&\scalebox{0.95}{$50.3$}&\scalebox{0.95}{$50.7$} &\scalebox{0.95}{$50.3$}&\scalebox{0.95}{$50.3$}&\cellcolor{lightroyalblue}\scalebox{0.95}{$50.5$}\\
    \scalebox{0.95}{FreDect \cite{frank2020leveraging}} &\scalebox{0.95}{$47.8$}&\scalebox{0.95}{$59.7$}&\scalebox{0.95}{$47.2$} &\scalebox{0.95}{$48.7$}&\scalebox{0.95}{$59.2$}&\cellcolor{lightroyalblue}\scalebox{0.95}{$52.5$}\\
    \scalebox{0.95}{Fusing \cite{ju2022fusing}} &\scalebox{0.95}{$50.0$}&\scalebox{0.95}{$50.0$}&\scalebox{0.95}{$50.5$} &\scalebox{0.95}{$50.1$}&\scalebox{0.95}{$49.9$}&\cellcolor{lightroyalblue}\scalebox{0.95}{$50.1$}\\
    \scalebox{0.95}{LGrad \cite{tan2023learning}} &\scalebox{0.95}{$54.7$}&\scalebox{0.95}{$44.5$}&\scalebox{0.95}{$44.4$} &\scalebox{0.95}{$44.8$}&\scalebox{0.95}{$46.5$}&\cellcolor{lightroyalblue}\scalebox{0.95}{$47.0$}\\
   \scalebox{0.95}{LNP \cite{liu2022detecting}} &\scalebox{0.95}{$58.2$}&\scalebox{0.95}{$41.3$}& \scalebox{0.95}{$43.1$}&\scalebox{0.95}{$53.3$}&\scalebox{0.95}{$42.5$}&\cellcolor{lightroyalblue}\scalebox{0.95}{$47.7$}\\
    \scalebox{0.95}{DIRE \cite{wang2023dire}} &\scalebox{0.95}{$50.2$}&\scalebox{0.95}{$49.8$}&\scalebox{0.95}{$50.3$} &\scalebox{0.95}{$50.1$}&\scalebox{0.95}{$50.6$}&\cellcolor{lightroyalblue}\scalebox{0.95}{$50.2$}\\
    \scalebox{0.95}{UnivFD \cite{ojha2023towards}} &\scalebox{0.95}{$48.5$}&\scalebox{0.95}{$52.0$}& \scalebox{0.95}{$53.4$}&\scalebox{0.95}{$60.9$}&\scalebox{0.95}{$50.7$}&\cellcolor{lightroyalblue}\scalebox{0.95}{$53.1$}\\
    
    \hline
  \end{tabularx}
  
  \vspace*{1mm}
  \hspace*{-1.5mm}
  \begin{tabularx}{\hsize}{|>{\raggedleft\arraybackslash}p{1.7cm}||Y|Y|Y|Y|Y|Y|}
    \hline\thickhline
   \rowcolor{mygray}\multirow{1}{*}{\scalebox{0.95}{mAP ($\%$)}}&\multicolumn{1}{c|}{\scalebox{0.95}{Pika}} 
     & \multicolumn{1}{c|}{\scalebox{0.95}{SVD}}& \multicolumn{1}{c|}{\scalebox{0.95}{OpS}}& \multicolumn{1}{c|}{\scalebox{0.95}{IXL}} &\multicolumn{1}{c|}{\scalebox{0.95}{Cog}} &\multicolumn{1}{c|}{\scalebox{0.95}{Avg.}}\\ 
    \hline\hline
    \scalebox{0.95}{Blind Guess} &\scalebox{0.95}{$50.0$}&\scalebox{0.95}{$50.0$}& \scalebox{0.95}{$50.0$}&\scalebox{0.95}{$50.0$}&\scalebox{0.95}{$50.0$}&\cellcolor{lightroyalblue}\scalebox{0.95}{$50.0$}\\\hline\hline
    \scalebox{0.95}{\scalebox{0.95}{CNNSpot} \cite{wang2020cnn}} &\scalebox{0.95}{$49.0$}&\scalebox{0.95}{$48.7$}&\scalebox{0.95}{$54.0$} &\scalebox{0.95}{$49.0$}&\scalebox{0.95}{$44.7$}&\cellcolor{lightroyalblue}\scalebox{0.95}{$49.1$}\\
    \scalebox{0.95}{FreDect \cite{frank2020leveraging}} &\scalebox{0.95}{$44.7$}&\scalebox{0.95}{$59.2$}&\scalebox{0.95}{$50.2$} &\scalebox{0.95}{$46.5$}&\scalebox{0.95}{$59.8$}&\cellcolor{lightroyalblue}\scalebox{0.95}{$52.1$}\\
    \scalebox{0.95}{Fusing \cite{ju2022fusing}} &\scalebox{0.95}{$47.7$}&\scalebox{0.95}{$47.6$}&\scalebox{0.95}{$60.1$} &\scalebox{0.95}{$58.7$}&\scalebox{0.95}{$44.3$}&\cellcolor{lightroyalblue}\scalebox{0.95}{$51.7$}\\
    \scalebox{0.95}{LGrad \cite{tan2023learning}} &\scalebox{0.95}{$56.8$}&\scalebox{0.95}{$43.0$}&\scalebox{0.95}{$42.2$} &\scalebox{0.95}{$43.2$}&\scalebox{0.95}{$45.2$}&\cellcolor{lightroyalblue}\scalebox{0.95}{$46.1$}\\
    \scalebox{0.95}{LNP \cite{liu2022detecting}} &\scalebox{0.95}{$82.1$}&\scalebox{0.95}{$38.8$}& \scalebox{0.95}{$37.3$}&\scalebox{0.95}{$72.3$}&\scalebox{0.95}{$41.9$}&\cellcolor{lightroyalblue}\scalebox{0.95}{$54.5$}\\
    \scalebox{0.95}{DIRE \cite{wang2023dire}} &\scalebox{0.95}{$49.8$}&\scalebox{0.95}{$49.4$}&\scalebox{0.95}{$47.9$} &\scalebox{0.95}{$49.0$}&\scalebox{0.95}{$51.5$}&\cellcolor{lightroyalblue}\scalebox{0.95}{$49.5$}\\
    \scalebox{0.95}{UnivFD \cite{ojha2023towards}} &\scalebox{0.95}{$40.1$}&\scalebox{0.95}{$56.2$}& \scalebox{0.95}{$60.1$}&\scalebox{0.95}{$72.6$}&\scalebox{0.95}{$48.6$}&\cellcolor{lightroyalblue}\scalebox{0.95}{$55.5$}\\
    
    \hline 
  \end{tabularx}
  \label{Table: ID}
  \vspace*{-4mm}
  \end{minipage}
\end{table}

\textbf{$\bullet$ The targeted training for poorly performing subjects.} The existing benchmarks, due to their limited subject scope, fail to inform researchers about the areas where their models perform well and where they fall short. With the TIP-Eval, researchers can identify underperforming subjects: for instance, in the case of the latest $\mathtt{CogVideoX\text{-}5B}$ \cite{yang2024cogvideox}, while it achieves an average \textit{aesthetic quality} of $0.74$ on the \textit{cottage} subject, it only reaches an average \textit{aesthetic quality} of $0.40$ on the \textit{calligraphy} subject. After identifying, researchers may gather targeted training videos and fine-tune their image-to-video models accordingly. 

\textbf{$\bullet$ Evaluating models' performance from \textit{direction} perspective.} While the existing benchmarks and our TIP-Eval evaluate image-to-video models across various subjects, it is equally important to assess whether the spatial transformations in the generated videos well-align with users’ expected \textbf{directions}.
As illustrated at the bottom of Fig. \ref{Fig: sort} and \ref{Fig: ratio}, our \dname~ encompasses a wide range of directions, such as ``\textit{zoom}", ``\textit{run}", and ``\textit{wave}", which are desired by users. Therefore, using our \dname~, future researchers may develop a benchmark specifically focused on directional control to complement existing ones.

\subsection{Identifying Generated Videos from Images}
\hspace{1.2em}Researchers should not overlook the safety issues in image-to-video generation while improving video quality. A key safety concern is \textbf{misinformation}, because image-to-video models can easily manipulate people or objects in images to make them perform actions they never actually did, as exemplified in Fig. \ref{Fig: fight}. In this section, we show how the proposed \dname~ helps combat such misinformation from the perspective of identifying generated videos from images. Specifically, the generated videos from five state-of-the-art models are split into separate sets to form \textbf{TIP-ID} dataset for training and testing the detectors. The details are shown in the Supplementary (Section \ref{Supple: ID}).


\begin{table}[t]
\centering
\begin{minipage}{0.485\textwidth}
\captionsetup{font=small}
\caption{Our trained strong detector’s performance in \textbf{classifying} videos as real, text-generated, or image-generated. `Same/Cross Domain’ refers to training and testing on the same or different diffusion models, respectively.} 
\vspace*{-2mm}
\hspace*{-1.5mm}
\small
  \begin{tabularx}{\hsize}{|>{\raggedleft\arraybackslash}p{1.8cm}||Y|Y|Y|Y|Y|Y|}
    \hline\thickhline
   \rowcolor{mygray}\multirow{1}{*}{\scalebox{0.95}{Accuracy ($\%$)}}&\multicolumn{1}{c|}{\scalebox{0.95}{Pika}} 
     & \multicolumn{1}{c|}{\scalebox{0.95}{SVD}}& \multicolumn{1}{c|}{\scalebox{0.95}{OpS}}& \multicolumn{1}{c|}{\scalebox{0.95}{IXL}} &\multicolumn{1}{c|}{\scalebox{0.95}{Cog}} &\multicolumn{1}{c|}{\scalebox{0.95}{Avg.}}\\ 
    \hline\hline
    \scalebox{0.95}{Blind Guess} &\scalebox{0.95}{$33.3$}&\scalebox{0.95}{$33.3$}& \scalebox{0.95}{$33.3$}&\scalebox{0.95}{$33.3$}&\scalebox{0.95}{$33.3$}&\cellcolor{lightroyalblue}\scalebox{0.95}{$33.3$}\\\hline\hline
    \scalebox{0.95}{Same Domain} &\scalebox{0.95}{$93.2$}&\scalebox{0.95}{$97.3$}&\scalebox{0.95}{$96.9$} &\scalebox{0.95}{$97.9$}&\scalebox{0.95}{$96.2$}&\cellcolor{lightroyalblue}\scalebox{0.95}{$96.3$}\\
    \scalebox{0.95}{Cross Domain} &\scalebox{0.95}{$84.5$}&\scalebox{0.95}{$92.1$}&\scalebox{0.95}{$93.4$} &\scalebox{0.95}{$73.6$}&\scalebox{0.95}{$92.2$}&\cellcolor{lightroyalblue}\scalebox{0.95}{$87.2$}\\
    \hline
  \end{tabularx}
  
  \vspace*{1mm}
  \hspace*{-1.5mm}
  \begin{tabularx}{\hsize}{|>{\raggedleft\arraybackslash}p{1.8cm}||Y|Y|Y|Y|Y|Y|}
    \hline\thickhline
   \rowcolor{mygray}\multirow{1}{*}{\scalebox{0.95}{mAP ($\%$)}}&\multicolumn{1}{c|}{\scalebox{0.95}{Pika}} 
     & \multicolumn{1}{c|}{\scalebox{0.95}{SVD}}& \multicolumn{1}{c|}{\scalebox{0.95}{OpS}}& \multicolumn{1}{c|}{\scalebox{0.95}{IXL}} &\multicolumn{1}{c|}{\scalebox{0.95}{Cog}} &\multicolumn{1}{c|}{\scalebox{0.95}{Avg.}}\\ 
    \hline\hline
    \scalebox{0.95}{Blind Guess} &\scalebox{0.95}{$33.3$}&\scalebox{0.95}{$33.3$}& \scalebox{0.95}{$33.3$}&\scalebox{0.95}{$33.3$}&\scalebox{0.95}{$33.3$}&\cellcolor{lightroyalblue}\scalebox{0.95}{$33.3$}\\\hline\hline
    \scalebox{0.95}{Same Domain} &\scalebox{0.95}{$98.7$}&\scalebox{0.95}{$99.7$}&\scalebox{0.95}{$99.6$} &\scalebox{0.95}{$99.8$}&\scalebox{0.95}{$99.4$}&\cellcolor{lightroyalblue}\scalebox{0.95}{$99.4$}\\
    \scalebox{0.95}{Cross Domain} &\scalebox{0.95}{$94.4$}&\scalebox{0.95}{$97.3$}&\scalebox{0.95}{$98.2$} &\scalebox{0.95}{$86.5$}&\scalebox{0.95}{$97.5$}&\cellcolor{lightroyalblue}\scalebox{0.95}{$94.8$}\\
    
    \hline
  \end{tabularx}
  \label{Table: ID_Train}
  \vspace*{-5mm}
  \end{minipage}
\end{table}

\textbf{A unique challenge in detecting generated videos from images.} 
As shown in Table \ref{Table: ID}, current fake image detection algorithms struggle to generalize when identifying such videos (note that because none of the state-of-the-art models can process entire videos directly, we use the middle frame from each video as the input image for each model.). This is because each frame in these videos can be considered a variant of the input real image, leading the detection algorithms to mistakenly classify these frames as real images. This unique characteristic of videos generated from images invalidates existing methods and calls for new efforts to address this challenge.

\textbf{A surprising and strong detector built by us.}
To address this challenge and establish a baseline for future research, we fine-tune a VideoMAE \cite{tong2022videomae} (see details in the Supplementary (Section \ref{Supple: VideoMAE})) to classify videos into three categories: (1) real, (2) text-generated, and (3) image-generated. The experiments are conducted in two settings: (1) \textit{Same domain}: we train and test the model on videos generated by the same diffusion model; for instance, both the training and testing videos are from $\mathtt{Pika}$. (2) \textit{Cross domain}: we train and test the model on videos generated by different diffusion models; for example, training videos are from $\mathtt{Stable}$ $\mathtt{Video}$ $\mathtt{Diffusion}$, $\mathtt{Open\text{-}Sora}$, $\mathtt{I2VGen\text{-}XL}$, and $\mathtt{CogVideoX\text{-}5B}$, while test videos are from $\mathtt{Pika}$. From the experimental results in Table \ref{Table: ID_Train}, we conclude that: \textbf{(1)} A simple classification model already achieves relative high performance in both same and cross domain settings. This result is somewhat surprising, as previous research, such as DIRE \cite{wang2023dire} and UnivFD \cite{ojha2023towards}, has shown that this naïve approach typically suffers from limited generalization. \textbf{(2)} There still remains a performance gap between the two settings (with a $9.1\%$ difference in accuracy and $4.6\%$ in mAP). Therefore, \textbf{future studies} may focus on enhancing the model’s generalizability to unseen diffusion models.

\subsection{Tracing the Source Images}

\begin{table}[t]
\centering
\begin{minipage}{0.485\textwidth}
\captionsetup{font=small}
\caption{The performance of publicly available pre-trained models and our trained models for \textbf{tracing} source images.} 
\vspace*{-2mm}
\hspace*{-2.2mm}
\small
\scalebox{1}{
  \begin{tabularx}{\hsize}{|>{\centering\arraybackslash}p{0.72cm}>{\raggedleft\arraybackslash}p{2cm}||Y|Y|Y|Y|Y|Y|}
    \hline\thickhline
    
   \rowcolor{mygray} \hspace*{-1mm}\scalebox{0.9}{$\mu$AP ($\%$)} &  \scalebox{0.9}{Method}&  \hspace*{-0.6mm}\scalebox{0.9}{Pika} & \hspace*{-1.0mm}\scalebox{0.9}{SVD} & \hspace*{-0.6mm}\scalebox{0.9}{Ops}&\hspace*{-0.5mm}\scalebox{0.9}{IXL}&\hspace*{-0.6mm}\scalebox{0.9}{Cog}& \hspace*{-0.7mm}\scalebox{0.9}{Avg.}\\ \hline \hline

    \multirow{3 }{*}{\vspace{-8.5mm}\rotatebox{90}{\shortstack{Supervised\\Pre-trained\\Models}}} & \scalebox{0.9}{Swin-B \citep{liu2021swin}}& \hspace*{-1.1mm}$74.4$& \hspace*{-1.1mm}$65.3$& \hspace*{-1.1mm}$25.9$& \hspace*{-1.1mm}$21.6$& \hspace*{-1.1mm}$67.7$& \cellcolor{lightroyalblue}\hspace*{-1.1mm}$51.0$ \\
    & \scalebox{0.9}{ResNet-50 \citep{he2016deep}} & \hspace*{-1.1mm}$80.7$& \hspace*{-1.1mm}$66.9$& \hspace*{-1.1mm}$23.4$& \hspace*{-1.1mm}$16.3$& \hspace*{-1.1mm}$72.4$& \cellcolor{lightroyalblue}\hspace*{-1.1mm}$51.9$ \\ 
     & \scalebox{0.9}{ConvNeXt \citep{liu2022convnet}}& \hspace*{-1.1mm}$77.1$& \hspace*{-1.1mm}$67.6$& \hspace*{-1.1mm}$26.1$& \hspace*{-1.1mm}$22.0$& \hspace*{-1.1mm}$69.4$& \cellcolor{lightroyalblue}\hspace*{-1.1mm}$52.4$ \\ 
     & \scalebox{0.9}{EfficientNet \citep{tan2019efficientnet}} & \hspace*{-1.1mm}$90.5$& \hspace*{-1.1mm}$79.5$& \hspace*{-1.1mm}$29.9$& \hspace*{-1.1mm}$24.0$& \hspace*{-1.1mm}$82.3$& \cellcolor{lightroyalblue}\hspace*{-1.1mm}$61.2$ \\
    & \scalebox{0.9}{ViT-B \citep{dosovitskiy2020vit}}& \hspace*{-1.1mm}$86.4$& \hspace*{-1.1mm}$74.6$& \hspace*{-1.1mm}$27.8$& \hspace*{-1.1mm}$21.7$& \hspace*{-1.1mm}$78.4$& \cellcolor{lightroyalblue}\hspace*{-1.1mm}$57.8$ \\

    \hline\hline
    \multirow{3 }{*}{\vspace{-8.6mm}\rotatebox{90}{\shortstack{\scalebox{0.9}{Self-supervised}\\Learning\\Models}}}& \scalebox{0.9}{SimSiam \citep{chen2021exploring}} & \hspace*{-1.1mm}$8.71$& \hspace*{-1.1mm}$5.13$& \hspace*{-1.1mm}$1.28$& \hspace*{-1.1mm}$0.74$& \hspace*{-1.1mm}$7.75$& \cellcolor{lightroyalblue}\hspace*{-1.1mm}$4.72$ \\
     & \scalebox{0.9}{MoCov3
     \citep{he2020momentum}} & \hspace*{-1.1mm}$32.0$& \hspace*{-1.1mm}$17.1$& \hspace*{-1.1mm}$1.97$& \hspace*{-1.1mm}$1.19$& \hspace*{-1.1mm}$26.3$& \cellcolor{lightroyalblue}\hspace*{-1.1mm}$15.7$\\
      & \scalebox{0.9}{DINOv2 \citep{oquab2023dinov2}} & \hspace*{-1.1mm}$73.1$& \hspace*{-1.1mm}$63.4$& \hspace*{-1.1mm}$25.6$& \hspace*{-1.1mm}$25.0$& \hspace*{-1.1mm}$66.4$& \cellcolor{lightroyalblue}\hspace*{-1.1mm}$50.7$\\ 
      & \scalebox{0.9}{MAE \citep{he2022masked}}  & \hspace*{-1.1mm}$37.2$& \hspace*{-1.1mm}$29.8$& \hspace*{-1.1mm}$2.02$& \hspace*{-1.1mm}$0.23$& \hspace*{-1.1mm}$28.6$& \cellcolor{lightroyalblue}\hspace*{-1.1mm}$19.6$\\
    & \scalebox{0.9}{SimCLR \citep{chen2020simple}} & \hspace*{-1.1mm}$93.3$& \hspace*{-1.1mm}$81.4$& \hspace*{-1.1mm}$21.2$& \hspace*{-1.1mm}$22.1$& \hspace*{-1.1mm}$86.2$& \cellcolor{lightroyalblue}\hspace*{-1.1mm}$60.8$ \\

    \hline\hline
     \multirow{2 }{*}{\vspace{-8.5mm}\rotatebox{90}{\shortstack{Vision-\\language\\Models}}}    & \scalebox{0.9}{CLIP \citep{radford2021learning}} & \hspace*{-1.1mm}$41.2$& \hspace*{-1.1mm}$28.1$& \hspace*{-1.1mm}$3.42$& \hspace*{-1.1mm}$1.85$& \hspace*{-1.1mm}$33.5$& \cellcolor{lightroyalblue}\hspace*{-1.1mm}$21.6$  \\ 
  & \scalebox{0.9}{SLIP \citep{mu2022slip}}  & \hspace*{-1.1mm}$82.0$& \hspace*{-1.1mm}$71.6$& \hspace*{-1.1mm}$26.3$& \hspace*{-1.1mm}$25.2$& \hspace*{-1.1mm}$75.8$& \cellcolor{lightroyalblue}\hspace*{-1.1mm}$56.2$ \\
   & \scalebox{0.9}{ZeroVL \citep{cui2022contrastive}} & \hspace*{-1.1mm}$68.4$& \hspace*{-1.1mm}$39.0$& \hspace*{-1.1mm}$5.41$& \hspace*{-1.1mm}$7.38$& \hspace*{-1.1mm}$51.8$& \cellcolor{lightroyalblue}\hspace*{-1.1mm}$34.4$\\ 
    & \scalebox{0.9}{BLIP \citep{li2022blip}}  & \hspace*{-1.1mm}$77.0$& \hspace*{-1.1mm}$68.9$& \hspace*{-1.1mm}$26.9$& \hspace*{-1.1mm}$23.7$& \hspace*{-1.1mm}$68.0$& \cellcolor{lightroyalblue}\hspace*{-1.1mm}$52.9$ \\

\hline\hline
    
     \multirow{3 }{*}{\vspace{-8.5mm}\rotatebox{90}{\shortstack{Image Copy\\Detection\\Models}}} & \scalebox{0.9}{ASL \citep{wang2023benchmark}}  & \hspace*{-1.1mm}$43.3$& \hspace*{-1.1mm}$31.4$& \hspace*{-1.1mm}$9.12$& \hspace*{-1.1mm}$5.91$& \hspace*{-1.1mm}$37.7$& \cellcolor{lightroyalblue}\hspace*{-1.1mm}$25.5$ \\ 
     & \scalebox{0.9}{CNNCL \citep{yokoo2021contrastive}} & \hspace*{-1.1mm}$93.0$& \hspace*{-1.1mm}$78.0$& \hspace*{-1.1mm}$27.6$& \hspace*{-1.1mm}$15.2$& \hspace*{-1.1mm}$83.5$& \cellcolor{lightroyalblue}\hspace*{-1.1mm}$59.5$\\
      & \scalebox{0.9}{BoT \citep{wang2021bag}} & \hspace*{-1.1mm}$98.3$& \hspace*{-1.1mm}$94.6$& \hspace*{-1.1mm}$47.1$& \hspace*{-1.1mm}$43.2$& \hspace*{-1.1mm}$94.0$& \cellcolor{lightroyalblue}\hspace*{-1.1mm}$75.4$ \\ 
    & \scalebox{0.9}{SSCD \citep{pizzi2022self}} & \hspace*{-1.1mm}$98.5$& \hspace*{-1.1mm}$95.5$& \hspace*{-1.1mm}$49.9$& \hspace*{-1.1mm}$47.9$& \hspace*{-1.1mm}$95.9$& \cellcolor{lightroyalblue}\hspace*{-1.1mm}$77.5$ \\
    & \scalebox{0.9}{AnyPattern \citep{wang2024AnyPattern}}& \hspace*{-1.1mm}$94.2$& \hspace*{-1.1mm}$89.5$& \hspace*{-1.1mm}$44.0$& \hspace*{-1.1mm}$48.3$& \hspace*{-1.1mm}$89.5$& \cellcolor{lightroyalblue}\hspace*{-1.1mm}$73.1$\\
    
    \hline\hline
    
     \multirow{-1 }{*}{\hspace*{2.5mm}\vspace{-5mm}\rotatebox{90}{\shortstack{Ours}}} & \scalebox{0.9}{Same Domain}  & \hspace*{-1.1mm}$99.1$& \hspace*{-1.1mm}$97.0$& \hspace*{-1.1mm}$61.5$& \hspace*{-1.1mm}$90.3$& \hspace*{-1.1mm}$97.6$& \cellcolor{lightroyalblue}\hspace*{-1.1mm}$89.1$ \\ 
     & \scalebox{0.9}{Cross Domain} & \hspace*{-1.1mm}$99.1$& \hspace*{-1.1mm}$96.9$& \hspace*{-1.1mm}$57.0$& \hspace*{-1.1mm}$73.3$& \hspace*{-1.1mm}$97.1$& \cellcolor{lightroyalblue}\hspace*{-1.1mm}$84.7$\\
       \hline 
  \end{tabularx}}
  \label{Table: plausible}
  \vspace*{-4mm}
  \end{minipage}
  \end{table}

\hspace{1.2em}In addition to identifying videos generated from images, in this section, we explore another approach to combat misinformation: given any frame from a generated video, we aim to retrieve its source image from a large database. For example, given a frame depicting `$\mathtt{Donald}$ $\mathtt{Trump}$ throwing a punch at $\mathtt{Elon}$ $\mathtt{Musk}$' (Fig. \ref{Fig: fight}, last frame), could we retrieve its source image showing `the two shaking hands friendly' (Fig. \ref{Fig: fight}, left)? If we achieve this, whenever malicious users attempt to mislead the public with any frame from generated videos, we can reveal the original source image to debunk the misinformation.
Based on \dname~, we construct a large database, \textbf{TIP-Trace}, with $4,590,000$ training, $90,000$ query, and $1,010,000$ reference images to conduct this study. For detailed dataset settings, please refer to the Supplementary (Section \ref{Supple: Trace}).

\textbf{The performance of existing pre-trained models is unsatisfactory for this task.} We benchmark existing methods on the TIP-Trace test set, including supervised pre-trained models, self-supervised learning models, vision-language models, and image copy detection models. From the Table \ref{Table: plausible}, we observe that the top-performing model, SSCD \cite{pizzi2022self}, achieves only $77.5\%$ $\mu$AP. This underscores the necessity of developing a specialized model for this task. 

\textbf{A strong baseline proposed by us.} We train a deep metric learning baseline aimed at creating a space where generated frames are close to their source images (see details in the Supplementary (Section \ref{Supple: metric})). During testing, we compare the cosine similarity between the query feature and each reference feature to identify the source image. Experiments in Table \ref{Table: plausible} show that, under the `Cross Domain' setting, the proposed baseline achieves a $+7.2\%$ superiority in $\mu$AP compared to its nearest competitor. However, we also observe that, compared to training and testing within the same diffusion model (`Same Domain'), there remains a significant performance gap for some diffusion models, such as $\mathtt{I2VGen\text{-}XL}$ \cite{zhang2023i2vgen}, with a decrease of $-17.0\%$ in $\mu$AP. Therefore, \textbf{future research} could focus on improving the generalizability of our baseline, making it more practical.

\subsection{Other Promising Research}
\hspace{1.2em}Beyond the research areas detailed above, we also introduce the following additional promising directions briefly.

\textbf{$\bullet$ Meaning-preserving prompt refinement.} 
Some user-provided text prompts in \textbf{\dname~} may be ambiguous and challenging for image-to-video models to interpret accurately. Based on our \textbf{\dname~}, future research could focus on designing prompt refiners to clarify these prompts while preserving the users’ original intent. 


\textbf{$\bullet$ Unsafe-prompt blocking.} Some prompts in \textbf{\dname~} may contain unsafe content, such as `two people fighting' or `asking a girl to undress'. Developing classifiers to detect and block such prompts before they are processed by image-to-video models is crucial for responsible AI deployment.

\textbf{$\bullet$ Copyright-respecting cartoon generation.}
About $11.0\%$ of image prompts in \textbf{\dname~} contain cartoon or animation elements, which may be subject to copyright. Future researchers may need to check whether the corresponding generated cartoon videos are similar to copyrighted real videos and try to prevent generating such content.

\textbf{$\bullet$ Cache for speeding-up generation.} When image-to-video models are matured and consistently produce high-quality videos, researchers could use these models with our prompts to pre-generate and cache videos. Then, when a new prompt is received, the video can be generated quickly based on the cache, potentially saving several minutes.

\section{Conclusion}
\hspace{1.2em}In this paper, we introduce \dname~, the first large-scale dataset of over $1.70$ million unique user-provided text and image prompts for image-to-video diffusion models. 
We compare \dname~ with existing prompt datasets, emphasizing the need for a specialized image-to-video prompt dataset due to differences in both basic and semantic content. 
Our dataset enables new research directions, such as better accommodating user preferences, developing more comprehensive and practical evaluation benchmarks, and addressing safety concerns like misinformation by identifying generated videos and tracing their source images. We encourage the research community to utilize and build upon our dataset to further advance the field.



\section*{Acknowledgment}
\hspace{1.2em}We sincerely thank OpenAI for their support through the Researcher Access Program. Without their generous contribution, this work would not have been possible.

{
    \small
    \bibliographystyle{ieeenat_fullname}
    \bibliography{main}
}

\clearpage
\setcounter{page}{1}

\maketitlesupplementary

\section{Comparing \dname~ with Panda-70M}
\label{sec:tvdataset}
\begin{figure}[t]
    \centering
    \includegraphics[width=0.48\textwidth]{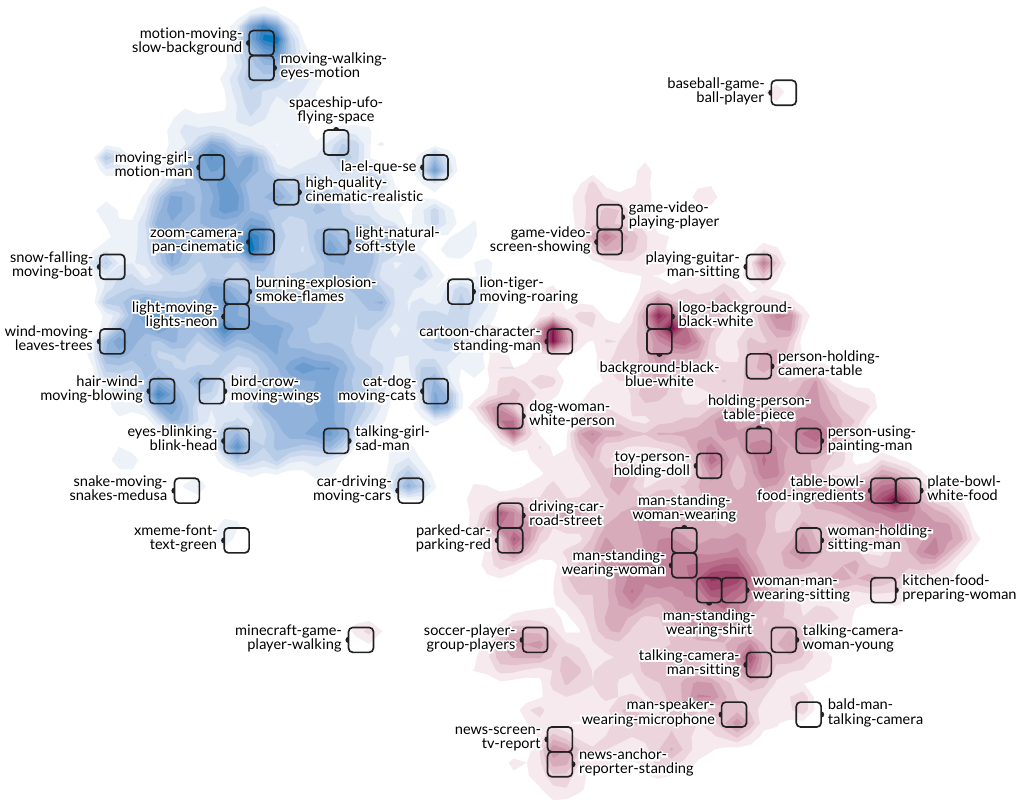}
    \vspace{-8mm}  
    \caption{The \textsc{WizMap} \cite{wang2023wizmap} visualization of our \textbf{\textcolor{bb}{\dname~}} compared to \textcolor{pp}{Panda-70M} \cite{chen2024panda}. Please \faSearch~zoom in to see the details.} 
    \label{Fig: panda}
   \vspace{-4mm} 
\end{figure}

\hspace{1.2em}As shown in Fig. \ref{Fig: panda}, we compare the text semantics of our \dname~ and Panda-70M \cite{chen2024panda} using \textsc{WizMap} \cite{wang2023wizmap} to highlight their differences.

\section{Exact Words from Pika’s Terms of Service}\label{Sup: License}

\begin{mdframed}
``\textbf{Your Inputs and Outputs.} You own all Outputs you create with the Service (``Your Outputs”). Notwithstanding the foregoing, nothing herein prevents Mellis or the Service from providing any Outputs to a third party that are the same as, or similar to, Your Outputs, and you hereby agree that such third party is free to use and exploit such Outputs without restriction from or obligation to you. You hereby grant Mellis and other users a license to any of your Inputs and Outputs that you make available to other users on the Service under the Creative Commons Noncommercial 4.0 Attribution International License (as accessible here: https://creativecommons.org/licenses/by-nc/4.0/legalcode)." --- \textit{Excerpt from Pika’s regulations}
\end{mdframed}

\section{Details of Adopted Image-to-Video Models}\label{Supple: i2v}

\hspace{1.2em}This section details the image-to-video models utilized in our \dname~ and the specifications we choose for each model, as shown in Table \ref{Table: Spec}.

$\mathtt{Pika}$ Image-to-Video \cite{pika2024} is a commercial AI-driven platform that transforms static images into dynamic video content. Users can sign up using a Discord account to access the platform’s services. Currently, the service is free to use; however, generated videos include a Pika Labs watermark and are intended for non-commercial purposes. Additionally, all created clips are publicly shared. 

$\mathtt{Stable}$ $\mathtt{Video}$ $\mathtt{Diffusion}$ \cite{blattmann2023stable} is an open-source generative AI model developed by Stability AI that transforms static images into short video clips (without text guidance). It is available in two versions: one generating $14$ frames and another producing $25$ frames, both supporting frame rates between $3$ and $30$ frames per second. 

$\mathtt{Open\text{-}Sora}$ \cite{opensora} is an open-source project developed by HPCAI Tech to democratize efficient video production. In Version $1.2$, it supports image-to-video generation for durations from $2$s to $15$s, resolutions from $144$p to $720$p, and any aspect ratio, effectively bringing the image to life.

$\mathtt{I2VGen\text{-}XL}$ \cite{zhang2023i2vgen}  is an advanced image-to-video synthesis model that generates high-quality videos from static images using a two-stage cascaded diffusion approach. To improve diversity, $\mathtt{I2VGen\text{-}XL}$ was trained on approximately $35$ million single-shot text-video pairs and $6$ billion text-image pairs. It addresses challenges in video synthesis like semantic accuracy, clarity, and spatio-temporal continuity.

$\mathtt{CogVideoX\text{-}5B}$ Image-to-Video \cite{yang2024cogvideox} is the latest AI model designed to generate dynamic videos from static images, guided by textual prompts. It is developed by the Knowledge Engineering Group at Tsinghua University and has 5 billion parameters.

\vspace{-1mm}
\begin{table}[t]
\centering
\begin{minipage}{0.485\textwidth}
\captionsetup{font=small}
\caption{The generated video specifications in our \dname~, including frame per second (FPS), duration, and resolution.} 
\vspace*{-2mm}
\small
  \begin{tabularx}{\hsize}{|>{\raggedleft\arraybackslash}p{3.7cm}||>{\centering\arraybackslash}p{0.35cm}|>{\centering\arraybackslash}p{0.45cm}|Y|}
    \hline\thickhline
   \rowcolor{mygray}\multirow{1}{*}{\scalebox{0.95}{Image-to-Video Models}}&\multicolumn{1}{c|}{\scalebox{0.95}{FPS}} 
     & \multicolumn{1}{c|}{\scalebox{0.95}{Duration}}& \multicolumn{1}{c|}{\scalebox{0.95}{Resolution}}\\ 
     \hline\hline
    \scalebox{0.95}{$\mathtt{Pika}$ \cite{pika2024}} &\scalebox{0.95}{\hspace{1mm}$24$}&\scalebox{0.95}{\hspace{4.5mm}$3$s}&\scalebox{0.95}{Varied}\\
    \scalebox{0.95}{\hspace*{-1mm}$\mathtt{Stable}$ $\mathtt{Video}$ $\mathtt{Diffusion}$ \cite{blattmann2023stable}} &\scalebox{0.95}{\hspace{1mm}$7$}&\scalebox{0.95}{\hspace{2.5mm}$3.57$s}&\scalebox{0.95}{\hspace*{-0.5mm}$1024 \times 576$}\\
    \scalebox{0.95}{$\mathtt{Open\text{-}Sora}$ \cite{opensora}}&\scalebox{0.95}{\hspace*{1mm}$24$}&\scalebox{0.95}{\hspace{2.5mm}$4.25$s}&\scalebox{0.95}{\hspace*{-0mm}$640 \times 360$}\\
    \scalebox{0.95}{$\mathtt{I2VGen\text{-}XL}$ \cite{zhang2023i2vgen}} &\scalebox{0.95}{\hspace{1mm}$7$}&\scalebox{0.95}{\hspace{4.5mm}$2$s}&\scalebox{0.95}{\hspace*{-0.5mm}$1280 \times 704$}\\
    \scalebox{0.95}{$\mathtt{CogVideoX\text{-}5B}$ \cite{yang2024cogvideox}} &\scalebox{0.95}{\hspace{1mm}$8$}&\scalebox{0.95}{\hspace{2.5mm}$6.13$s}&\scalebox{0.95}{\hspace*{-0mm}$720 \times 480$}\\
    
    \hline
  \end{tabularx}
  \label{Table: Spec}
  \vspace*{-5mm}
  \end{minipage}
\end{table}

\begin{figure*}[t]
    \centering
    \includegraphics[width=0.98\textwidth]{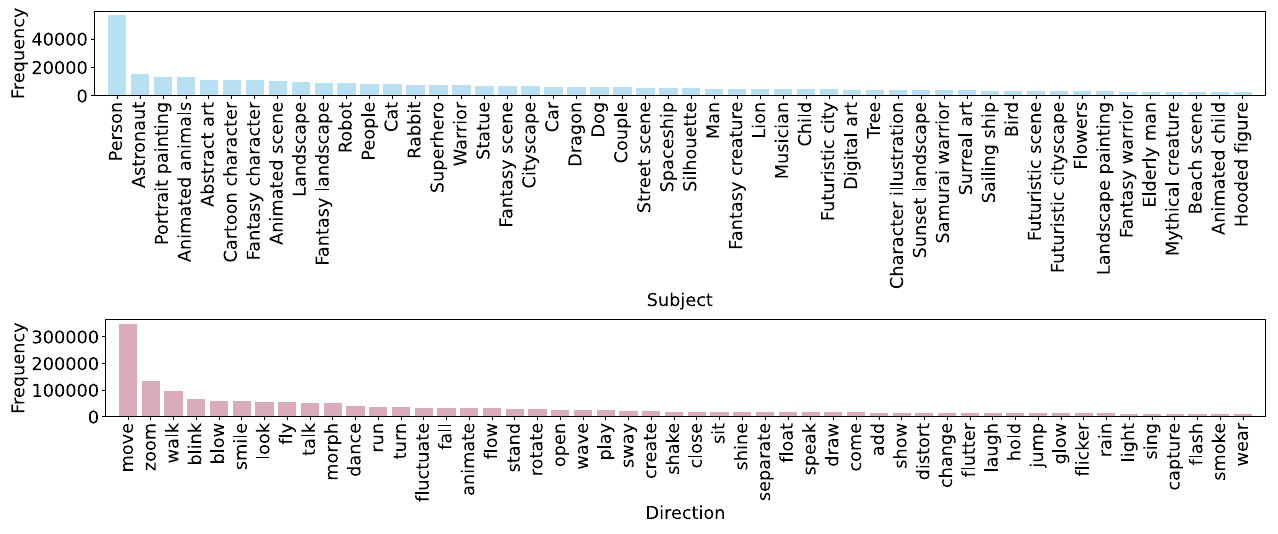}
    \vspace{-2mm}  
    \caption{An extension of Fig. \ref{Fig: sort}: the top $50$ \textit{subjects} (top) and \textit{directions} (bottom) preferred by users when generating videos from images.} 
    \label{Fig: sort_50}
   \vspace{-4mm} 
\end{figure*}

\section{Details of Calculating User Preference}
\label{Supple: User}

\hspace{1.2em}\textbf{Calculate the most popular subjects:} \textbf{(1)} for each data point, embed the subject using $\mathtt{SentenceTransformer}$ \cite{reimers2019sentence} to obtain a $384$-dimensional vector; \textbf{(2)} cluster the resulting $1,701,935$ vectors using $\mathtt{HDBSCAN}$ \cite{mcinnes2017hdbscan}, which automatically generates $21,247$ clusters; and \textbf{(3)} for each cluster, use the most frequent subject as the representative and then rank the obtained subjects by frequency. \textbf{Note that} we adopt this approach because $\mathtt{GPT}$-$\mathtt{4o}$ \cite{openai2024hello} may use slightly different variations for the same subject. For example, for the subject \textit{`Dragon'}, $\mathtt{GPT}$-$\mathtt{4o}$ \cite{openai2024hello} may output \textit{`Dragon'}, \textit{`Dragons'}, \textit{`Dragon, creature'} or \textit{`Dragon creature'}.

\textbf{Calculate the most popular directions:} \textbf{(1)} use $\mathtt{GPT}$-$\mathtt{4o}$ \cite{openai2024hello} to extract each verb from the text prompts; \textbf{(2)} gather all extracted verbs; and \textbf{(3)} rank them by frequency. The used prompt for $\mathtt{GPT}$-$\mathtt{4o}$ \cite{openai2024hello} is:
\begin{mdframed}
\textit{``Extract verbs in a given sentence, return their base form, separated by commas, and do not return anything else. If there is no verb, please return ` '. "}
\end{mdframed}

\section{Examples for Top Subjects and Directions}\label{Supple: Example}
\begin{figure*}[t]
    \centering
    \includegraphics[width=0.98\textwidth]{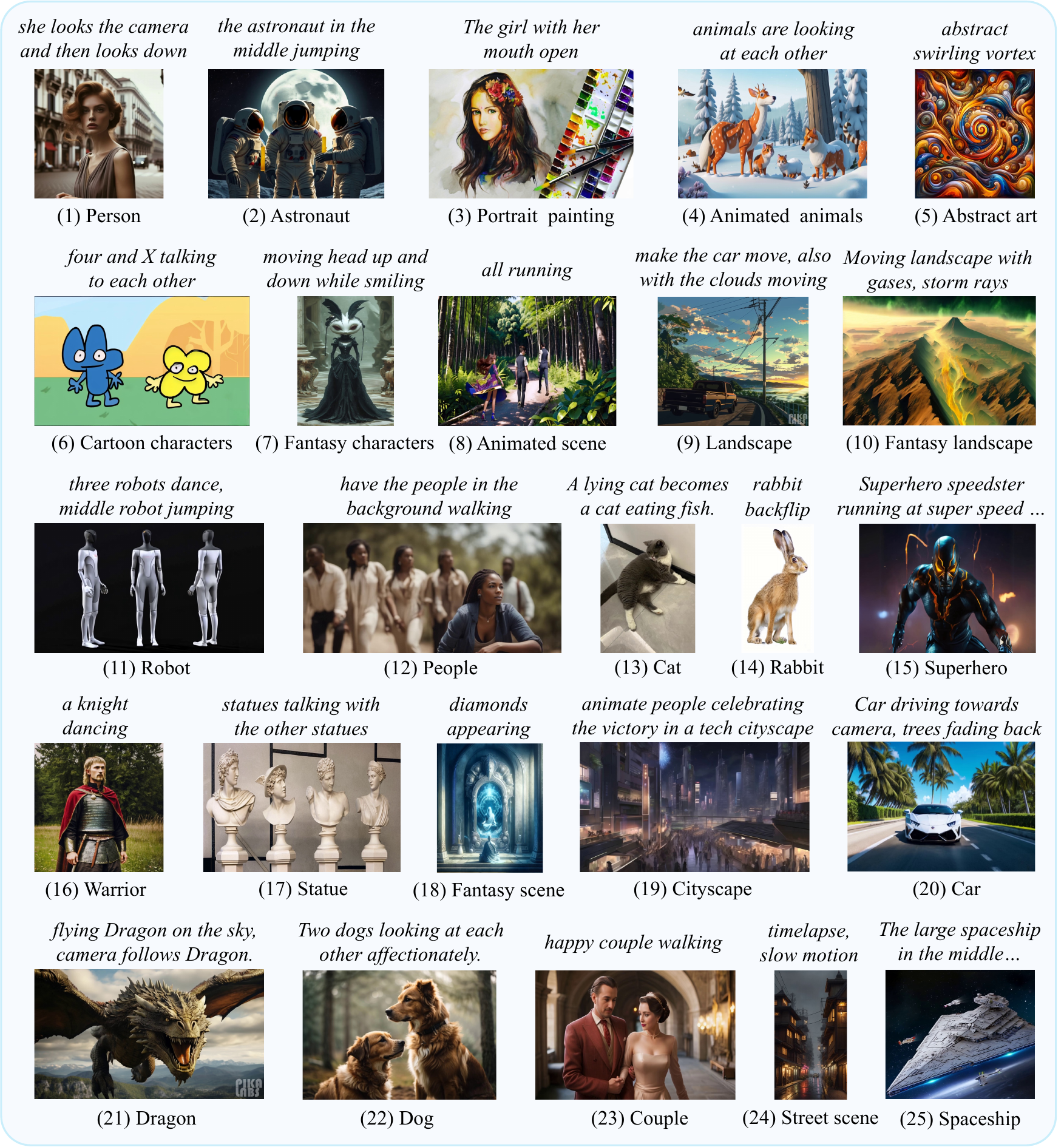}
    \vspace{-2mm}  
    \caption{For each top-ranked \textbf{subject}, we select one text and one image prompt as examples for illustration.} 
    \label{Fig: subject}
   \vspace{-5mm} 
\end{figure*}

\begin{figure*}[t]
    \centering
    \includegraphics[width=0.98\textwidth]{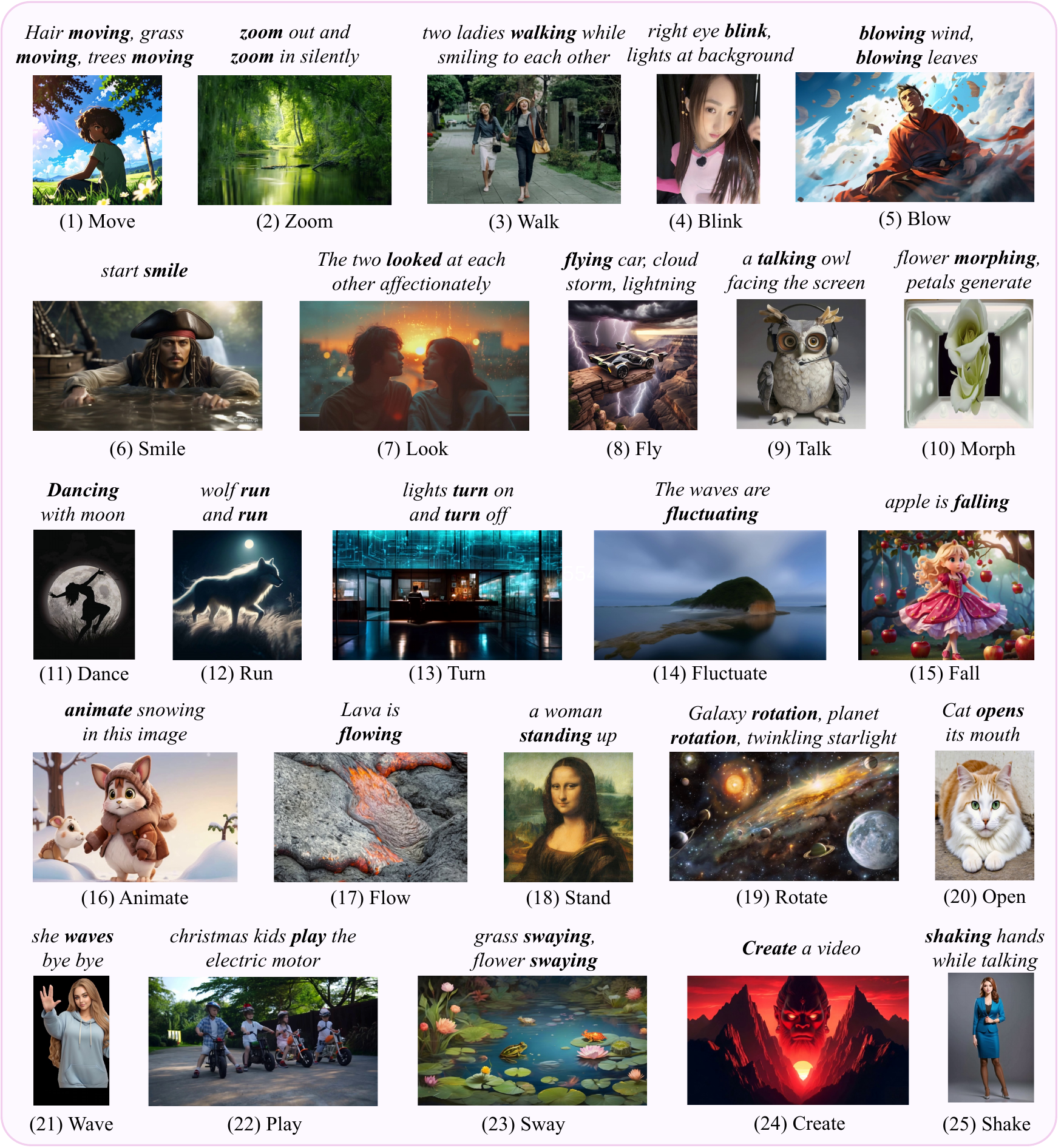}
    \vspace{-2mm}  
    \caption{For each top-ranked \textbf{direction}, we select one text and one image prompt as examples for illustration.} 
    \label{Fig: direction}
   \vspace{-5mm} 
\end{figure*}

\hspace{1.2em}As shown in Fig. \ref{Fig: subject} and Fig. \ref{Fig: direction}, for each of the top $25$ most popular subjects and directions, we select one text and one image prompt for illustration. Beyond this, in Fig. \ref{Fig: sort_50}, we extend Fig. \ref{Fig: sort} to show the top 50 users’ preferred \textit{subjects} (top) and \textit{directions} (bottom).

\section{Full Experiments for Benchmarking}\label{Supple: Bench}

\begin{table}[t]
\centering
\begin{minipage}{0.485\textwidth}
\captionsetup{font=small}
\caption{The full experimental results for drawing Fig. \ref{Fig: bench}. Similar to VBench \cite{huang2023vbench}, when drawing the radar chart, results are normalized per dimension to a common scale between $0.3$ and $0.8$ linearly.} 
\vspace*{-2mm}
\small
  \begin{tabularx}{\hsize}{|>{\raggedleft\arraybackslash}p{2.75cm}||Y|Y|Y|Y|Y|Y|}
    \hline\thickhline
   \rowcolor{mygray}\multirow{1}{*}{\scalebox{0.95}{Dimension}}&\multicolumn{1}{c|}{\scalebox{0.95}{Pika}} 
     & \multicolumn{1}{c|}{\scalebox{0.95}{SVD}}& \multicolumn{1}{c|}{\scalebox{0.95}{OpS}}& \multicolumn{1}{c|}{\scalebox{0.95}{IXL}} &\multicolumn{1}{c|}{\scalebox{0.95}{Cog}}\\ 
     \hline\hline
    \scalebox{0.95}{Subject Consistency} &\scalebox{0.95}{\hspace*{-0.5mm}$0.976$}&\scalebox{0.95}{\hspace*{-0.5mm}$0.950$}&\scalebox{0.95}{\hspace*{-0.5mm}$0.826$} &\scalebox{0.95}{\hspace*{-0.5mm}$0.816$}&\scalebox{0.95}{\hspace*{-0.5mm}$0.949$}\\
    \scalebox{0.92}{\hspace*{-1.6mm}Background Consistency} &\scalebox{0.95}{\hspace*{-0.5mm}$0.981$}&\scalebox{0.95}{\hspace*{-0.5mm}$0.959$}&\scalebox{0.95}{\hspace*{-0.5mm}$0.909$} &\scalebox{0.95}{\hspace*{-0.5mm}$0.893$}&\scalebox{0.95}{\hspace*{-0.5mm}$0.962$}\\
    \scalebox{0.95}{Motion Smoothness} &\scalebox{0.95}{\hspace*{-0.5mm}$0.995$}&\scalebox{0.95}{\hspace*{-0.5mm}$0.984$}&\scalebox{0.95}{\hspace*{-0.5mm}$0.992$} &\scalebox{0.95}{\hspace*{-0.5mm}$0.948$}&\scalebox{0.95}{\hspace*{-0.5mm}$0.983$}\\
    \scalebox{0.95}{Dynamic Degree} &\scalebox{0.95}{\hspace*{-0.5mm}$0.058$}&\scalebox{0.95}{\hspace*{-0.5mm}$0.667$}&\scalebox{0.95}{\hspace*{-0.5mm}$0.326$} &\scalebox{0.95}{\hspace*{-0.5mm}$0.775$}&\scalebox{0.95}{\hspace*{-0.5mm}$0.253$}\\
    \scalebox{0.95}{Aesthetic Quality} &\scalebox{0.95}{\hspace*{-0.5mm}$0.659$}&\scalebox{0.95}{\hspace*{-0.5mm}$0.585$}&\scalebox{0.95}{\hspace*{-0.5mm}$0.512$} &\scalebox{0.95}{\hspace*{-0.5mm}$0.555$}&\scalebox{0.95}{\hspace*{-0.5mm}$0.611$}\\
    \scalebox{0.95}{Imaging Quality} &\scalebox{0.95}{\hspace*{-0.5mm}$0.627$}&\scalebox{0.95}{\hspace*{-0.5mm}$0.586$}&\scalebox{0.95}{\hspace*{-0.5mm}$0.514$} &\scalebox{0.95}{\hspace*{-0.5mm}$0.551$}&\scalebox{0.95}{\hspace*{-0.5mm}$0.617$}\\
    \scalebox{0.95}{Temporal Consistency} &\scalebox{0.95}{\hspace*{-0.5mm}$0.997$}&\scalebox{0.95}{\hspace*{-0.5mm}$0.984$}&\scalebox{0.95}{\hspace*{-0.5mm}$0.987$} &\scalebox{0.95}{\hspace*{-0.5mm}$0.953$}&\scalebox{0.95}{\hspace*{-0.5mm}$0.989$}\\
    \scalebox{0.95}{Video-text Alignment} &\scalebox{0.95}{\hspace*{-0.5mm}$0.254$}&\scalebox{0.95}{\hspace*{-0.5mm}$0.252$}&\scalebox{0.95}{\hspace*{-0.5mm}$0.258$} &\scalebox{0.95}{\hspace*{-0.5mm}$0.260$}&\scalebox{0.95}{\hspace*{-0.5mm}$0.255$}\\
    \scalebox{0.95}{\hspace*{-1.6mm}Video-image Alignment} &\scalebox{0.95}{\hspace*{-0.5mm}$0.974$}&\scalebox{0.95}{\hspace*{-0.5mm}$0.932$}&\scalebox{0.95}{\hspace*{-0.5mm}$0.767$} &\scalebox{0.95}{\hspace*{-0.5mm}$0.791$}&\scalebox{0.95}{\hspace*{-0.5mm}$0.946$}\\
    \scalebox{0.95}{DOVER \cite{wu2023dover}} &\scalebox{0.95}{\hspace*{-0.5mm}$0.713$}&\scalebox{0.95}{\hspace*{-0.5mm}$0.607$}&\scalebox{0.95}{\hspace*{-0.5mm}$0.460$} &\scalebox{0.95}{\hspace*{-0.5mm}$0.478$}&\scalebox{0.95}{\hspace*{-0.5mm}$0.652$}\\

    \hline
  \end{tabularx}
  \label{Table: Radar_full}
  \vspace*{-4mm}
  \end{minipage}
\end{table}

\hspace{1.2em}Table \ref{Table: Radar_full} provides the full experiments for generating the radar chart shown in Fig. \ref{Fig: bench}. For the selected $10$ dimensions, `\textit{subject consistency}', `\textit{background consistency}’, `\textit{motion smoothness}', `\textit{dynamic degree}’, `\textit{aesthetic quality}', and `\textit{imaging quality}’ are derived from $\mathtt{VBench}$-$\mathtt{I2V}$ \cite{huang2023vbench} and $\mathtt{I2V}$-$\mathtt{Bench}$ \cite{renconsisti2v}, while `\textit{temporal consistency}', `\textit{video-text alignment}', `\textit{video-image alignment}', and `\textit{disentangled objective video quality evaluator (DOVER)}' \cite{wu2023dover} are from $\mathtt{AIGCBench}$ \cite{fan2024aigcbench}.

\section{Details of TIP-ID Dataset}
\label{Supple: ID}

\hspace{1.2em}Unlike previous fake image detection datasets, which classify images into \textbf{two} classes -- real and fake -- our TIP-ID dataset emphasizes \textbf{three} classes: real videos, videos generated from texts, and videos generated from images.

\textbf{Sources.} \textbf{(1) Real videos.} The real videos are sourced from the VSC22 dataset \cite{pizzi20242023}, which comprises approximately $100,000$ videos derived from the YFCC100M dataset \cite{thomee2016yfcc100m}, ensuring diversity and comprehensiveness. To match the lengths of generated videos, we split the real videos into $3$-second segments. This results in \underline{$354,486$} real videos totally. \textbf{(2) Videos generated from texts.} We randomly select \underline{$400,000$} text-generated videos from VidProM \cite{wang2024vidprom}, with $100,000$ videos from each text-to-video diffusion model: $\mathtt{Pika}$ \cite{pika2024}, $\mathtt{VideoCraft2}$ \cite{chen2024videocrafter2}, $\mathtt{Text2Video}$-$\mathtt{Zero}$ \cite{khachatryan2023text2video}, and $\mathtt{ModelScope}$ \cite{wang2023modelscope}. \textbf{(3) Videos generated from images.} We use \underline{$500,000$} image-generated videos in our \dname~, with $100,000$ videos from each image-to-video diffusion model. With these sources, the constructed TIP-ID dataset is relatively balanced across each class.

\textbf{Split.} We split the TIP-ID dataset into a 9:1 ratio for training and testing, respectively. It is important to note that: \textbf{(1)} When benchmarking existing fake image detection methods, we exclude the class of text-generated videos, as these methods can only classify videos (images) as real or fake. \textbf{(2)} For the training and test sets of image-generated videos, UUIDs do not overlap. 
This restriction is intended to prevent potential data leakage, as for the same UUID, diffusion models generate videos from the same image. \textbf{(3)} Although we split real videos into segments, the segments of any given real video are assigned to either the training set or the test set, but not both. This is also for preventing potential data leakage.

\textbf{Settings.} We consider two settings for evaluating detectors on our TIP-ID dataset. \textbf{(1) Same domain.} Both the training and testing image-generated videos are generated by the same diffusion model. For instance, we train and test a detector on videos generated by $\mathtt{Open}$-$\mathtt{Sora}$. This setting aims to test whether a detector can achieve high performance when it has already encountered videos generated by one diffusion model, which can be considered the upper bound for the next setting.
\textbf{(2) Cross domain.} The training and testing image-generated videos are generated by different diffusion models. For example, we train a detector on videos generated by $\mathtt{Pika}$, $\mathtt{Stable\ Video\ Diffusion}$, $\mathtt{I2VGen\text{-}XL}$, and $\mathtt{CogVideoX\text{-}5B}$, but test it on $\mathtt{Open\text{-}Sora}$. This approach is more practical, as a trained detector will likely encounter newly-developed image-to-video models that it has not previously seen.

\begin{figure*}[t]
    \centering
    \includegraphics[width=0.98\textwidth]{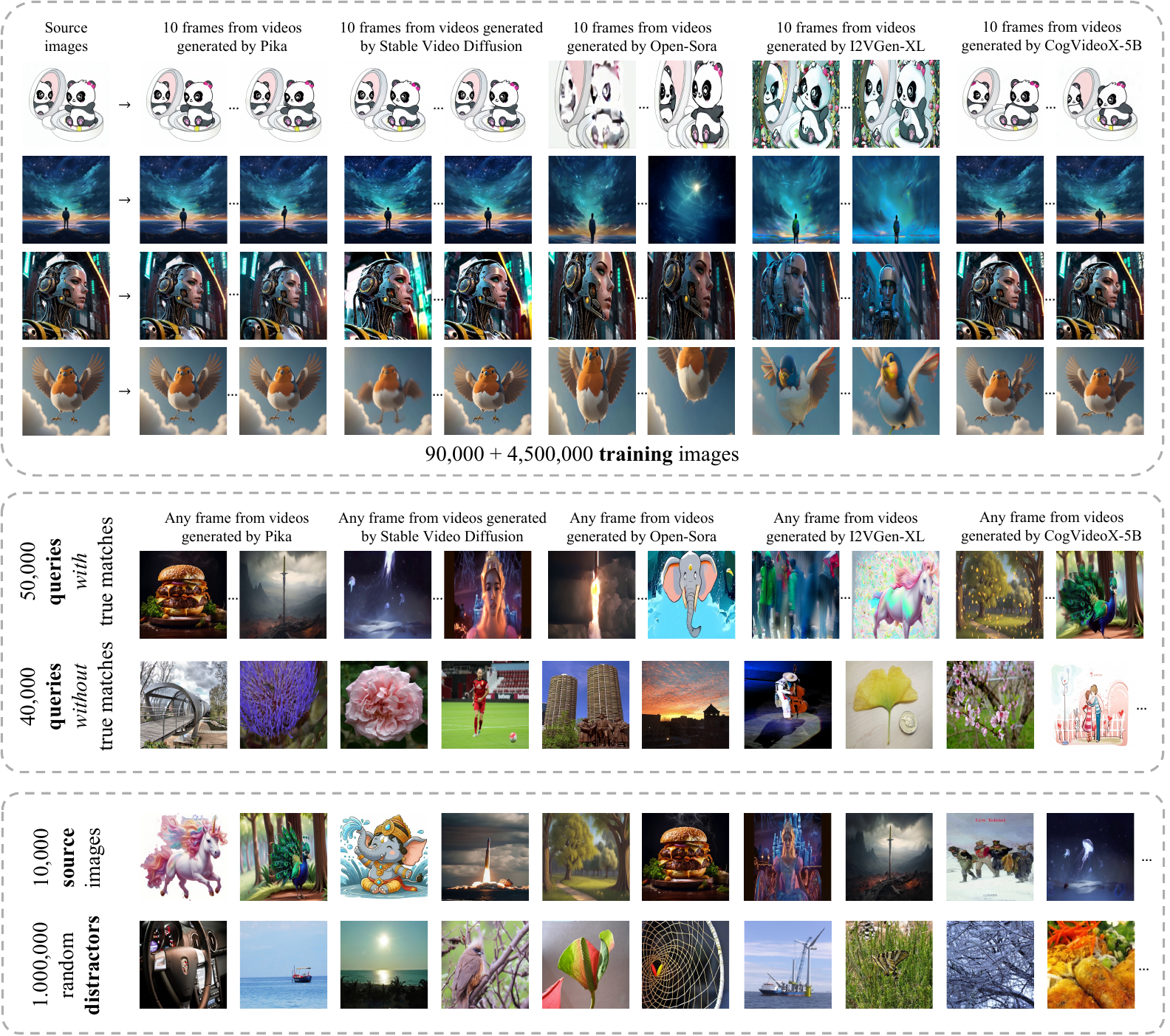}
    \vspace{-2mm}  
    \caption{The illustration of the \textbf{TIP-Trace}, which is designed to train a model to identify the source image of any given generated frame.} 
    \label{Fig: tip_trace}
   \vspace{-5mm} 
\end{figure*}
\textbf{Evaluation metrics.} Following the fake image detection task, we use Accuracy and Mean Average Precision (mAP) to evaluate the performance of models on the proposed TIP-ID dataset. Specifically, Accuracy measures the proportion of correct predictions among all predictions; whereas mAP evaluates performance for each class, which is useful for handling class imbalances.

\section{Details of Fine-tuning VideoMAE}\label{Supple: VideoMAE}
\hspace{1.2em}We fine-tune the Video Masked Autoencoder (VideoMAE) \cite{tong2022videomae} on the TIP-ID dataset for a video classification task. Specifically, our preprocessing pipeline includes (1) temporal subsampling, (2) spatial transformations, and (3) normalization. During training, the spatial transformations consist of random short-side scaling, random cropping to $224 \times 224$, and random horizontal flipping; for testing, we only resize frames to $224 \times 224$. The pre-trained model is downloaded from the VideoMAE's official Hugging Face
repository, and we adjust it to match the number of classes in our dataset, \ie, 3, by updating the classification head. Training is distributed across a server with $8$ A100 GPUs. The mini-batch size is $8$, the learning rate is $5\times 10^{-5}$, and the number of iterations is $20,000$.

\section{Details of TIP-Trace Dataset}
\label{Supple: Trace}
\hspace{1.2em}As shown in Fig. \ref{Fig: tip_trace}, this section provides a detailed description of the proposed TIP-Trace dataset. It includes \textit{training}, \textit{query}, and \textit{reference} sets:

\textbf{$\bullet$ Training set.} Recall that we randomly selected $100,000$ text and image prompts from \dname~ to generate videos using state-of-the-art image-to-video models. We use $90,000$ of these text and image prompts to construct the training set. Specifically, for the videos generated by each image-to-video model, we uniformly select $10$ frames from each, resulting in a total of $5 \times 90,000 \times 10 = 4,500,000$ training images. Including the image prompts (source images), we have a total of  $90,000 + 4,500,000 = 4,590,000$ training images, as shown in Fig. \ref{Fig: tip_trace} (top).

\textbf{$\bullet$ Query set.} As shown in Fig. \ref{Fig: tip_trace} (middle), the query set consists of two parts: \textbf{(1) Queries generated from remaining 10,000 prompts.} 	Instead of uniformly selecting $10$ frames from each video, we randomly pick one frame from each video for testing. This results in $50,000$ query images with true matches. \textbf{(2) Distractor queries.} The distractor images, \ie, images not extracted from image-to-video generation, serve to replicate real-world scenarios where there is an abundance of authentic images rather than artificially generated ones. We randomly select $40,000$ from Open Images Dataset \cite{kuznetsova2020open} as the distractor queries.
 
\textbf{$\bullet$ Reference set.} We design the reference set to mimic a ``needle-in-a-haystack" scenario in the real world, where the majority of images do not have corresponding queries. Specifically, as shown in Fig. \ref{Fig: tip_trace} (bottom), we incorporate the $10,000$ source images into a set of $1,000,000$ reference images from DISC21 \cite{papakipos2022results}, which is derived from the real-world multimedia dataset YFCC100M \cite{thomee2016yfcc100m}.

Beyond the split sets, we also introduce two evaluation settings and one evaluation metric to assess model performance on the proposed dataset:

\textbf{$\bullet$ Two evaluation settings.} We consider two settings for evaluating model performance on the TIP-Trace dataset. \textbf{(1) Same domain.} In this setting, models can be trained and tested on data from all $5$ image-to-video models. This setting is used to evaluate whether a model can learn discriminative information after training. \textbf{(2) Cross domain.} We observe that, in the real world, new image-to-video models continually emerge. Therefore, in this setting, we assess whether a trained model can generalize to unseen models. Specifically, we exclude one of the five models from the training set and conduct testing on the excluded model. This setting is more challenging and practical than the first.

\textbf{$\bullet$ An evaluation metric.} 
This task uses $\mu$AP (micro Average Precision) as the evaluation metric. $\mu$AP considers the overall performance across all queries by aggregating true positives, false positives, and false negatives over the entire dataset before calculating precision and recall. This evaluation metric is particularly suitable for this task as it provides a more holistic measure of a model’s effectiveness in distinguishing between matching and non-matching images in large-scale datasets.

\section{Details of Deep Metric Learning Baseline}\label{Supple: metric}
\hspace{1.2em}We first treat each source image and its generated frames together as a single class, then \textbf{train} a CosFace \cite{wang2018cosface} on the resulting $90,000$ classes as a strong deep metric learning baseline. The hyperparameters are set as follows: the model architecture is ViT-Base \cite{dosovitskiy2021an}, with a CosFace loss margin of $0.35$ and a scale parameter of $64$. The training process is with a batch size of $512$, using $4$ instances per class. We use a cosine learning rate schedule with a maximum learning rate of $0.00035$ and a warmup period of $5$ epochs. The model is trained for $25$ epochs, with $2,000$ iterations per epoch, distributed across $8$ A100 GPUs. Input images are resized to a height and width of $224 \times 224$. When \textbf{testing}, we remove the classification layer and use the trained ViT-Base to extract features from queries and references.

\section{Potential Social Impact}
\hspace{1.2em}The \dname~ dataset has potential for \textbf{positive} social impact by \textit{enhancing digital creativity} and \textit{fostering responsible AI use}. Specifically, by helping the creation of more user-responsive image-to-video models, TIP-I2V enables content creators to create engaging and customized videos. Additionally, \dname~ contributes to the development of detection models that help verify authenticity, trace image sources, and prevent harmful content misuse. Nevertheless, the \dname~ dataset may also have potential \textbf{negative} social impacts if \textbf{misused}. 
Below, we outline several potential negative social impacts and provide solutions:

\textbf{$\bullet$ NSFW content.} Although limited in quantity, our dataset includes some NSFW text and image prompts, which may be sensitive or potentially discomforting for certain individuals. Similar to VidProM \cite{wang2024vidprom} and DiffusionDB \cite{wang2023diffusiondb}, we choose not to remove these NSFW prompts, as they may provide valuable data for AI safety researchers to analyze and develop content-blocking solutions. Nevertheless, we provide NSFW scores for text and image prompts, allowing regular researchers to easily identify and remove these prompts if they find the content uncomfortable.

\textbf{$\bullet$ Privacy.} Although, per Pika’s regulations, users agree to make their input and output publicly available, some may still feel uncomfortable with the inclusion of their prompts in \dname~. To enhance user privacy, we implement the following measures: (1) each prompt is assigned a new UUID and an anonymous UserID instead of the identifiable original user name; and (2) users have the right to request that their contributions be removed from \dname~. They can simply email us to make this request.

\textbf{$\bullet$ Copyright.} According to Pika Labs’ Terms of Service, users are responsible for ensuring that their content does not violate any copyright laws or third-party rights. However, we have noticed that Pika Labs lacks preventive measures, leading some users to upload images, such as ``Mickey Mouse", which may be subject to copyright restrictions. Nevertheless, including these images in our \dname~ does \textbf{not} constitute copyright infringement, as our usage falls under \textit{fair use}. While our dataset is open-sourced under a non-commercial license (CC BY-NC 4.0), some malicious users may still use this dataset for commercial purposes, potentially infringing copyright. Therefore, we strongly recommend that users of \dname~ comply with our license to avoid any legal risks.


\end{document}


%% file: main.bbl
\begin{thebibliography}{74}
\providecommand{\natexlab}[1]{#1}
\providecommand{\url}[1]{\texttt{#1}}
\expandafter\ifx\csname urlstyle\endcsname\relax
  \providecommand{\doi}[1]{doi: #1}\else
  \providecommand{\doi}{doi: \begingroup \urlstyle{rm}\Url}\fi

\bibitem[hai(2024)]{hailuo_ai_video}
Hailuo ai video generator - reimagine video creation - image-to-video, 2024.

\bibitem[kua(2024)]{kuaishou_kling}
Kling - kuaishou - image-to-video, 2024.

\bibitem[met(2024)]{meta2024moviegen}
Movie gen: Ai video generation tool, 2024.

\bibitem[ope(2024)]{openai2024sora}
Sora: Ai text-to-video model, 2024.

\bibitem[pik(2024)]{pika2024}
Pika - creative video editing platform, 2024.

\bibitem[Bach et~al.(2022)Bach, Sanh, Yong, Webson, Raffel, Nayak, Sharma, Kim,
  Bari, Fevry, et~al.]{bach2022promptsource}
Stephen Bach, Victor Sanh, Zheng~Xin Yong, Albert Webson, Colin Raffel, Nihal~V
  Nayak, Abheesht Sharma, Taewoon Kim, M~Saiful Bari, Thibault Fevry, et~al.
\newblock Promptsource: An integrated development environment and repository
  for natural language prompts.
\newblock \emph{Annual Meeting of the Association for Computational
  Linguistics}, 2022.

\bibitem[Bain et~al.(2021)Bain, Nagrani, Varol, and Zisserman]{bain2021frozen}
Max Bain, Arsha Nagrani, G{\"u}l Varol, and Andrew Zisserman.
\newblock Frozen in time: A joint video and image encoder for end-to-end
  retrieval.
\newblock \emph{IEEE/CVF International Conference on Computer Vision}, 2021.

\bibitem[Blattmann et~al.(2023)Blattmann, Dockhorn, Kulal, Mendelevitch,
  Kilian, Lorenz, Levi, English, Voleti, Letts, et~al.]{blattmann2023stable}
Andreas Blattmann, Tim Dockhorn, Sumith Kulal, Daniel Mendelevitch, Maciej
  Kilian, Dominik Lorenz, Yam Levi, Zion English, Vikram Voleti, Adam Letts,
  et~al.
\newblock Stable video diffusion: Scaling latent video diffusion models to
  large datasets.
\newblock \emph{arXiv preprint arXiv:2311.15127}, 2023.

\bibitem[Chen et~al.(2023)Chen, Xia, He, Zhang, Cun, Yang, Xing, Liu, Chen,
  Wang, et~al.]{chen2023videocrafter1}
Haoxin Chen, Menghan Xia, Yingqing He, Yong Zhang, Xiaodong Cun, Shaoshu Yang,
  Jinbo Xing, Yaofang Liu, Qifeng Chen, Xintao Wang, et~al.
\newblock Videocrafter1: Open diffusion models for high-quality video
  generation.
\newblock \emph{arXiv preprint arXiv:2310.19512}, 2023.

\bibitem[Chen et~al.(2024{\natexlab{a}})Chen, Zhang, Cun, Xia, Wang, Weng, and
  Shan]{chen2024videocrafter2}
Haoxin Chen, Yong Zhang, Xiaodong Cun, Menghan Xia, Xintao Wang, Chao Weng, and
  Ying Shan.
\newblock Videocrafter2: Overcoming data limitations for high-quality video
  diffusion models.
\newblock \emph{IEEE/CVF Conference on Computer Vision and Pattern
  Recognition}, 2024{\natexlab{a}}.

\bibitem[Chen et~al.(2020)Chen, Kornblith, Norouzi, and Hinton]{chen2020simple}
Ting Chen, Simon Kornblith, Mohammad Norouzi, and Geoffrey Hinton.
\newblock A simple framework for contrastive learning of visual
  representations.
\newblock \emph{International conference on machine learning}, 2020.

\bibitem[Chen et~al.(2024{\natexlab{b}})Chen, Siarohin, Menapace, Deyneka,
  Chao, Jeon, Fang, Lee, Ren, Yang, et~al.]{chen2024panda}
Tsai-Shien Chen, Aliaksandr Siarohin, Willi Menapace, Ekaterina Deyneka,
  Hsiang-wei Chao, Byung~Eun Jeon, Yuwei Fang, Hsin-Ying Lee, Jian Ren,
  Ming-Hsuan Yang, et~al.
\newblock Panda-70m: Captioning 70m videos with multiple cross-modality
  teachers.
\newblock \emph{IEEE/CVF Conference on Computer Vision and Pattern
  Recognition}, 2024{\natexlab{b}}.

\bibitem[Chen and He(2021)]{chen2021exploring}
Xinlei Chen and Kaiming He.
\newblock Exploring simple siamese representation learning.
\newblock \emph{IEEE/CVF conference on computer vision and pattern
  recognition}, 2021.

\bibitem[Cui et~al.(2022)Cui, Zhou, Guo, Yin, Wu, Yoshie, and
  Chen]{cui2022contrastive}
Quan Cui, Boyan Zhou, Yu Guo, Weidong Yin, Hao Wu, Osamu Yoshie, and Yubo Chen.
\newblock Contrastive vision-language pre-training with limited resources.
\newblock \emph{European Conference on Computer Vision}, 2022.

\bibitem[Dosovitskiy et~al.(2021{\natexlab{a}})Dosovitskiy, Beyer, Kolesnikov,
  Weissenborn, Zhai, Unterthiner, Dehghani, Minderer, Heigold, Gelly,
  Uszkoreit, and Houlsby]{dosovitskiy2020vit}
Alexey Dosovitskiy, Lucas Beyer, Alexander Kolesnikov, Dirk Weissenborn,
  Xiaohua Zhai, Thomas Unterthiner, Mostafa Dehghani, Matthias Minderer, Georg
  Heigold, Sylvain Gelly, Jakob Uszkoreit, and Neil Houlsby.
\newblock An image is worth 16x16 words: Transformers for image recognition at
  scale.
\newblock \emph{International Conference on Learning Representations},
  2021{\natexlab{a}}.

\bibitem[Dosovitskiy et~al.(2021{\natexlab{b}})Dosovitskiy, Beyer, Kolesnikov,
  Weissenborn, Zhai, Unterthiner, Dehghani, Minderer, Heigold, Gelly,
  Uszkoreit, and Houlsby]{dosovitskiy2021an}
Alexey Dosovitskiy, Lucas Beyer, Alexander Kolesnikov, Dirk Weissenborn,
  Xiaohua Zhai, Thomas Unterthiner, Mostafa Dehghani, Matthias Minderer, Georg
  Heigold, Sylvain Gelly, Jakob Uszkoreit, and Neil Houlsby.
\newblock An image is worth 16x16 words: Transformers for image recognition at
  scale.
\newblock \emph{International Conference on Learning Representations},
  2021{\natexlab{b}}.

\bibitem[{Falconsai}(2024)]{huggingface2024nsfw}
{Falconsai}.
\newblock Falconsai: nsfw image detection, 2024.

\bibitem[Fan et~al.(2024)Fan, Luo, Gao, and Zhan]{fan2024aigcbench}
Fanda Fan, Chunjie Luo, Wanling Gao, and Jianfeng Zhan.
\newblock Aigcbench: Comprehensive evaluation of image-to-video content
  generated by ai.
\newblock \emph{BenchCouncil Transactions on Benchmarks, Standards and
  Evaluations}, 2024.

\bibitem[Frank et~al.(2020)Frank, Eisenhofer, Sch{\"o}nherr, Fischer, Kolossa,
  and Holz]{frank2020leveraging}
Joel Frank, Thorsten Eisenhofer, Lea Sch{\"o}nherr, Asja Fischer, Dorothea
  Kolossa, and Thorsten Holz.
\newblock Leveraging frequency analysis for deep fake image recognition.
\newblock \emph{International conference on machine learning}, 2020.

\bibitem[Golub(2024)]{DiscordChatExporter}
Alexey Golub.
\newblock Discordchatexporter, 2024.

\bibitem[Guo et~al.(2024)Guo, Zheng, Hou, Gao, Deng, Wan, Zhang, Liu, Hu, Zha,
  et~al.]{guo2024i2v}
Xun Guo, Mingwu Zheng, Liang Hou, Yuan Gao, Yufan Deng, Pengfei Wan, Di Zhang,
  Yufan Liu, Weiming Hu, Zhengjun Zha, et~al.
\newblock I2v-adapter: A general image-to-video adapter for diffusion models.
\newblock \emph{ACM SIGGRAPH 2024 Conference Papers}, 2024.

\bibitem[He et~al.(2016)He, Zhang, Ren, and Sun]{he2016deep}
Kaiming He, Xiangyu Zhang, Shaoqing Ren, and Jian Sun.
\newblock Deep residual learning for image recognition.
\newblock \emph{IEEE conference on computer vision and pattern recognition},
  2016.

\bibitem[He et~al.(2020)He, Fan, Wu, Xie, and Girshick]{he2020momentum}
Kaiming He, Haoqi Fan, Yuxin Wu, Saining Xie, and Ross Girshick.
\newblock Momentum contrast for unsupervised visual representation learning.
\newblock \emph{IEEE/CVF conference on computer vision and pattern
  recognition}, 2020.

\bibitem[He et~al.(2022)He, Chen, Xie, Li, Doll{\'a}r, and
  Girshick]{he2022masked}
Kaiming He, Xinlei Chen, Saining Xie, Yanghao Li, Piotr Doll{\'a}r, and Ross
  Girshick.
\newblock Masked autoencoders are scalable vision learners.
\newblock \emph{IEEE/CVF conference on computer vision and pattern
  recognition}, 2022.

\bibitem[Huang et~al.(2024)Huang, He, Yu, Zhang, Si, Jiang, Zhang, Wu, Jin,
  Chanpaisit, Wang, Chen, Wang, Lin, Qiao, and Liu]{huang2023vbench}
Ziqi Huang, Yinan He, Jiashuo Yu, Fan Zhang, Chenyang Si, Yuming Jiang, Yuanhan
  Zhang, Tianxing Wu, Qingyang Jin, Nattapol Chanpaisit, Yaohui Wang, Xinyuan
  Chen, Limin Wang, Dahua Lin, Yu Qiao, and Ziwei Liu.
\newblock {VBench}: Comprehensive benchmark suite for video generative models.
\newblock \emph{IEEE/CVF Conference on Computer Vision and Pattern
  Recognition}, 2024.

\bibitem[Ju et~al.(2024)Ju, Gao, Zhang, Yuan, Wang, Zeng, Xiong, Xu, and
  Shan]{ju2024miradata}
Xuan Ju, Yiming Gao, Zhaoyang Zhang, Ziyang Yuan, Xintao Wang, Ailing Zeng, Yu
  Xiong, Qiang Xu, and Ying Shan.
\newblock Miradata: A large-scale video dataset with long durations and
  structured captions.
\newblock \emph{Thirty-eighth Conference on Neural Information Processing
  Systems}, 2024.

\bibitem[Ju et~al.(2022)Ju, Jia, Ke, Xue, Nagano, and Lyu]{ju2022fusing}
Yan Ju, Shan Jia, Lipeng Ke, Hongfei Xue, Koki Nagano, and Siwei Lyu.
\newblock Fusing global and local features for generalized ai-synthesized image
  detection.
\newblock \emph{IEEE International Conference on Image Processing}, 2022.

\bibitem[Khachatryan et~al.(2023{\natexlab{a}})Khachatryan, Movsisyan,
  Tadevosyan, Henschel, Wang, Navasardyan, and Shi]{khachatryan2023text2video}
Levon Khachatryan, Andranik Movsisyan, Vahram Tadevosyan, Roberto Henschel,
  Zhangyang Wang, Shant Navasardyan, and Humphrey Shi.
\newblock Text2video-zero: Text-to-image diffusion models are zero-shot video
  generators.
\newblock \emph{IEEE/CVF International Conference on Computer Vision},
  2023{\natexlab{a}}.

\bibitem[Khachatryan et~al.(2023{\natexlab{b}})Khachatryan, Movsisyan,
  Tadevosyan, Henschel, Wang, Navasardyan, and Shi]{text2video-zero}
Levon Khachatryan, Andranik Movsisyan, Vahram Tadevosyan, Roberto Henschel,
  Zhangyang Wang, Shant Navasardyan, and Humphrey Shi.
\newblock Text2video-zero: Text-to-image diffusion models are zero-shot video
  generators.
\newblock \emph{IEEE/CVF International Conference on Computer Vision},
  2023{\natexlab{b}}.

\bibitem[Kuznetsova et~al.(2020)Kuznetsova, Rom, Alldrin, Uijlings, Krasin,
  Pont-Tuset, Kamali, Popov, Malloci, Kolesnikov, et~al.]{kuznetsova2020open}
Alina Kuznetsova, Hassan Rom, Neil Alldrin, Jasper Uijlings, Ivan Krasin, Jordi
  Pont-Tuset, Shahab Kamali, Stefan Popov, Matteo Malloci, Alexander
  Kolesnikov, et~al.
\newblock The open images dataset v4: Unified image classification, object
  detection, and visual relationship detection at scale.
\newblock \emph{International journal of computer vision}, 2020.

\bibitem[Lab and etc.(2024)]{pku_yuan_lab}
PKU-Yuan Lab and Tuzhan~AI etc.
\newblock Open-sora-plan, 2024.

\bibitem[Li et~al.(2022)Li, Li, Xiong, and Hoi]{li2022blip}
Junnan Li, Dongxu Li, Caiming Xiong, and Steven Hoi.
\newblock Blip: Bootstrapping language-image pre-training for unified
  vision-language understanding and generation.
\newblock \emph{International Conference on Machine Learning}, 2022.

\bibitem[Liu et~al.(2022{\natexlab{a}})Liu, Yang, Bi, Xiao, Li, and
  Gao]{liu2022detecting}
Bo Liu, Fan Yang, Xiuli Bi, Bin Xiao, Weisheng Li, and Xinbo Gao.
\newblock Detecting generated images by real images.
\newblock \emph{European Conference on Computer Vision}, 2022{\natexlab{a}}.

\bibitem[Liu et~al.(2021)Liu, Lin, Cao, Hu, Wei, Zhang, Lin, and
  Guo]{liu2021swin}
Ze Liu, Yutong Lin, Yue Cao, Han Hu, Yixuan Wei, Zheng Zhang, Stephen Lin, and
  Baining Guo.
\newblock Swin transformer: Hierarchical vision transformer using shifted
  windows.
\newblock \emph{IEEE/CVF international conference on computer vision}, 2021.

\bibitem[Liu et~al.(2022{\natexlab{b}})Liu, Mao, Wu, Feichtenhofer, Darrell,
  and Xie]{liu2022convnet}
Zhuang Liu, Hanzi Mao, Chao-Yuan Wu, Christoph Feichtenhofer, Trevor Darrell,
  and Saining Xie.
\newblock A convnet for the 2020s.
\newblock \emph{IEEE/CVF conference on computer vision and pattern
  recognition}, 2022{\natexlab{b}}.

\bibitem[McInnes et~al.(2017)McInnes, Healy, and Astels]{mcinnes2017hdbscan}
Leland McInnes, John Healy, and Steve Astels.
\newblock Hdbscan: Hierarchical density based clustering.
\newblock \emph{The Journal of Open Source Software}, 2017.

\bibitem[Mu et~al.(2022)Mu, Kirillov, Wagner, and Xie]{mu2022slip}
Norman Mu, Alexander Kirillov, David Wagner, and Saining Xie.
\newblock Slip: Self-supervision meets language-image pre-training.
\newblock \emph{European Conference on Computer Vision}, 2022.

\bibitem[Nan et~al.(2024)Nan, Xie, Zhou, Fan, Yang, Chen, Li, Yang, and
  Tai]{nan2024openvid}
Kepan Nan, Rui Xie, Penghao Zhou, Tiehan Fan, Zhenheng Yang, Zhijie Chen, Xiang
  Li, Jian Yang, and Ying Tai.
\newblock Openvid-1m: A large-scale high-quality dataset for text-to-video
  generation.
\newblock \emph{arXiv preprint arXiv:2407.02371}, 2024.

\bibitem[Ni et~al.(2023)Ni, Shi, Li, Huang, and Min]{ni2023conditional}
Haomiao Ni, Changhao Shi, Kai Li, Sharon~X Huang, and Martin~Renqiang Min.
\newblock Conditional image-to-video generation with latent flow diffusion
  models.
\newblock \emph{IEEE/CVF conference on computer vision and pattern
  recognition}, 2023.

\bibitem[Ojha et~al.(2023)Ojha, Li, and Lee]{ojha2023towards}
Utkarsh Ojha, Yuheng Li, and Yong~Jae Lee.
\newblock Towards universal fake image detectors that generalize across
  generative models.
\newblock \emph{IEEE/CVF Conference on Computer Vision and Pattern
  Recognition}, 2023.

\bibitem[{OpenAI}(2024{\natexlab{a}})]{openai2024embedding}
{OpenAI}.
\newblock New embedding models and api updates, 2024{\natexlab{a}}.

\bibitem[{OpenAI}(2024{\natexlab{b}})]{openai2024hello}
{OpenAI}.
\newblock Hello gpt-4o, 2024{\natexlab{b}}.

\bibitem[Oquab et~al.(2023)Oquab, Darcet, Moutakanni, Vo, Szafraniec, Khalidov,
  Fernandez, Haziza, Massa, El-Nouby, Howes, Huang, Xu, Sharma, Li, Galuba,
  Rabbat, Assran, Ballas, Synnaeve, Misra, Jegou, Mairal, Labatut, Joulin, and
  Bojanowski]{oquab2023dinov2}
Maxime Oquab, Timothée Darcet, Theo Moutakanni, Huy~V. Vo, Marc Szafraniec,
  Vasil Khalidov, Pierre Fernandez, Daniel Haziza, Francisco Massa, Alaaeldin
  El-Nouby, Russell Howes, Po-Yao Huang, Hu Xu, Vasu Sharma, Shang-Wen Li,
  Wojciech Galuba, Mike Rabbat, Mido Assran, Nicolas Ballas, Gabriel Synnaeve,
  Ishan Misra, Herve Jegou, Julien Mairal, Patrick Labatut, Armand Joulin, and
  Piotr Bojanowski.
\newblock Dinov2: Learning robust visual features without supervision.
\newblock \emph{Transactions on Machine Learning Research}, 2023.

\bibitem[Papakipos et~al.(2022)Papakipos, Tolias, Jenicek, Pizzi, Yokoo, Wang,
  Sun, Zhang, Yang, Addicam, et~al.]{papakipos2022results}
Zo{\"e} Papakipos, Giorgos Tolias, Tomas Jenicek, Ed Pizzi, Shuhei Yokoo,
  Wenhao Wang, Yifan Sun, Weipu Zhang, Yi Yang, Sanjay Addicam, et~al.
\newblock Results and findings of the 2021 image similarity challenge.
\newblock \emph{NeurIPS 2021 Competitions and Demonstrations Track}, 2022.

\bibitem[Pizzi et~al.(2022)Pizzi, Roy, Ravindra, Goyal, and
  Douze]{pizzi2022self}
Ed Pizzi, Sreya~Dutta Roy, Sugosh~Nagavara Ravindra, Priya Goyal, and Matthijs
  Douze.
\newblock A self-supervised descriptor for image copy detection.
\newblock \emph{IEEE/CVF Conference on Computer Vision and Pattern
  Recognition}, 2022.

\bibitem[Pizzi et~al.(2024)Pizzi, Kordopatis-Zilos, Patel, Postelnicu,
  Ravindra, Gupta, Papadopoulos, Tolias, and Douze]{pizzi20242023}
Ed Pizzi, Giorgos Kordopatis-Zilos, Hiral Patel, Gheorghe Postelnicu,
  Sugosh~Nagavara Ravindra, Akshay Gupta, Symeon Papadopoulos, Giorgos Tolias,
  and Matthijs Douze.
\newblock The 2023 video similarity dataset and challenge.
\newblock \emph{Computer Vision and Image Understanding}, 2024.

\bibitem[Radford et~al.(2021)Radford, Kim, Hallacy, Ramesh, Goh, Agarwal,
  Sastry, Askell, Mishkin, Clark, et~al.]{radford2021learning}
Alec Radford, Jong~Wook Kim, Chris Hallacy, Aditya Ramesh, Gabriel Goh,
  Sandhini Agarwal, Girish Sastry, Amanda Askell, Pamela Mishkin, Jack Clark,
  et~al.
\newblock Learning transferable visual models from natural language
  supervision.
\newblock \emph{International conference on machine learning}, 2021.

\bibitem[Reimers and Gurevych(2019)]{reimers2019sentence}
Nils Reimers and Iryna Gurevych.
\newblock Sentence-bert: Sentence embeddings using siamese bert-networks.
\newblock \emph{Proceedings of the 2019 Conference on Empirical Methods in
  Natural Language Processing and the 9th International Joint Conference on
  Natural Language Processing (EMNLP-IJCNLP)}, 2019.

\bibitem[Ren et~al.()Ren, Yang, Zhang, Wei, Du, Huang, and Chen]{renconsisti2v}
Weiming Ren, Huan Yang, Ge Zhang, Cong Wei, Xinrun Du, Wenhao Huang, and Wenhu
  Chen.
\newblock Consisti2v: Enhancing visual consistency for image-to-video
  generation.
\newblock \emph{Transactions on Machine Learning Research}.

\bibitem[Tan et~al.(2023)Tan, Zhao, Wei, Gu, and Wei]{tan2023learning}
Chuangchuang Tan, Yao Zhao, Shikui Wei, Guanghua Gu, and Yunchao Wei.
\newblock Learning on gradients: Generalized artifacts representation for
  gan-generated images detection.
\newblock \emph{IEEE/CVF Conference on Computer Vision and Pattern
  Recognition}, 2023.

\bibitem[Tan and Le(2019)]{tan2019efficientnet}
Mingxing Tan and Quoc Le.
\newblock Efficientnet: Rethinking model scaling for convolutional neural
  networks.
\newblock \emph{International conference on machine learning}, 2019.

\bibitem[Thomee et~al.(2016)Thomee, Shamma, Friedland, Elizalde, Ni, Poland,
  Borth, and Li]{thomee2016yfcc100m}
Bart Thomee, David~A Shamma, Gerald Friedland, Benjamin Elizalde, Karl Ni,
  Douglas Poland, Damian Borth, and Li-Jia Li.
\newblock Yfcc100m: The new data in multimedia research.
\newblock \emph{Communications of the ACM}, 59\penalty0 (2):\penalty0 64--73,
  2016.

\bibitem[Tong et~al.(2022)Tong, Song, Wang, and Wang]{tong2022videomae}
Zhan Tong, Yibing Song, Jue Wang, and Limin Wang.
\newblock Videomae: Masked autoencoders are data-efficient learners for
  self-supervised video pre-training.
\newblock \emph{Advances in neural information processing systems}, 2022.

\bibitem[{Unitary team}(2020)]{Detoxify}
{Unitary team}.
\newblock Detoxify, 2020.

\bibitem[Wang et~al.(2018)Wang, Wang, Zhou, Ji, Gong, Zhou, Li, and
  Liu]{wang2018cosface}
Hao Wang, Yitong Wang, Zheng Zhou, Xing Ji, Dihong Gong, Jingchao Zhou, Zhifeng
  Li, and Wei Liu.
\newblock Cosface: Large margin cosine loss for deep face recognition.
\newblock \emph{IEEE/CVF conference on computer vision and pattern
  recognition}, 2018.

\bibitem[Wang et~al.(2023{\natexlab{a}})Wang, Yuan, Chen, Zhang, Wang, and
  Zhang]{wang2023modelscope}
Jiuniu Wang, Hangjie Yuan, Dayou Chen, Yingya Zhang, Xiang Wang, and Shiwei
  Zhang.
\newblock Modelscope text-to-video technical report.
\newblock \emph{arXiv preprint arXiv:2308.06571}, 2023{\natexlab{a}}.

\bibitem[Wang et~al.(2020)Wang, Wang, Zhang, Owens, and Efros]{wang2020cnn}
Sheng-Yu Wang, Oliver Wang, Richard Zhang, Andrew Owens, and Alexei~A Efros.
\newblock \emph{IEEE/CVF conference on computer vision and pattern
  recognition}, 2020.

\bibitem[Wang and Yang(2024)]{wang2024vidprom}
Wenhao Wang and Yi Yang.
\newblock Vidprom: A million-scale real prompt-gallery dataset for
  text-to-video diffusion models.
\newblock \emph{Thirty-eighth Conference on Neural Information Processing
  Systems}, 2024.

\bibitem[Wang et~al.(2021)Wang, Zhang, Sun, and Yang]{wang2021bag}
Wenhao Wang, Weipu Zhang, Yifan Sun, and Yi Yang.
\newblock Bag of tricks and a strong baseline for image copy detection.
\newblock \emph{arXiv preprint arXiv:2111.08004}, 2021.

\bibitem[Wang et~al.(2023{\natexlab{b}})Wang, Sun, and Yang]{wang2023benchmark}
Wenhao Wang, Yifan Sun, and Yi Yang.
\newblock A benchmark and asymmetrical-similarity learning for practical image
  copy detection.
\newblock \emph{AAAI Conference on Artificial Intelligence},
  2023{\natexlab{b}}.

\bibitem[Wang et~al.(2024{\natexlab{a}})Wang, Sun, Tan, and
  Yang]{wang2024AnyPattern}
Wenhao Wang, Yifan Sun, Zhentao Tan, and Yi Yang.
\newblock Anypattern: Towards in-context image copy detection.
\newblock In \emph{arXiv preprint arXiv:2404.13788}, 2024{\natexlab{a}}.

\bibitem[Wang et~al.(2024{\natexlab{b}})Wang, He, Li, Li, Yu, Ma, Li, Chen,
  Chen, Wang, et~al.]{wanginternvid}
Yi Wang, Yinan He, Yizhuo Li, Kunchang Li, Jiashuo Yu, Xin Ma, Xinhao Li, Guo
  Chen, Xinyuan Chen, Yaohui Wang, et~al.
\newblock Internvid: A large-scale video-text dataset for multimodal
  understanding and generation.
\newblock \emph{The Twelfth International Conference on Learning
  Representations}, 2024{\natexlab{b}}.

\bibitem[Wang et~al.(2023{\natexlab{c}})Wang, Bao, Zhou, Wang, Hu, Chen, and
  Li]{wang2023dire}
Zhendong Wang, Jianmin Bao, Wengang Zhou, Weilun Wang, Hezhen Hu, Hong Chen,
  and Houqiang Li.
\newblock Dire for diffusion-generated image detection.
\newblock \emph{IEEE/CVF International Conference on Computer Vision},
  2023{\natexlab{c}}.

\bibitem[Wang et~al.(2023{\natexlab{d}})Wang, Montoya, Munechka, Yang, Hoover,
  and Chau]{wang2023diffusiondb}
Zijie Wang, Evan Montoya, David Munechka, Haoyang Yang, Benjamin Hoover, and
  Polo Chau.
\newblock Diffusiondb: A large-scale prompt gallery dataset for text-to-image
  generative models.
\newblock \emph{Annual Meeting of the Association for Computational
  Linguistics}, 2023{\natexlab{d}}.

\bibitem[Wang et~al.(2023{\natexlab{e}})Wang, Hohman, and Chau]{wang2023wizmap}
Zijie~J Wang, Fred Hohman, and Duen~Horng Chau.
\newblock Wizmap: Scalable interactive visualization for exploring large
  machine learning embeddings.
\newblock \emph{Annual Meeting Of The Association For Computational
  Linguistics}, 2023{\natexlab{e}}.

\bibitem[Wu et~al.(2023)Wu, Zhang, Liao, Chen, Hou, Wang, Sun, Yan, and
  Lin]{wu2023dover}
Haoning Wu, Erli Zhang, Liang Liao, Chaofeng Chen, Jingwen~Hou Hou, Annan Wang,
  Wenxiu~Sun Sun, Qiong Yan, and Weisi Lin.
\newblock Exploring video quality assessment on user generated contents from
  aesthetic and technical perspectives.
\newblock \emph{International Conference on Computer Vision}, 2023.

\bibitem[Xing et~al.(2023)Xing, Feng, Chen, Dai, Hu, Xu, Wu, and
  Jiang]{xing2023survey}
Zhen Xing, Qijun Feng, Haoran Chen, Qi Dai, Han Hu, Hang Xu, Zuxuan Wu, and
  Yu-Gang Jiang.
\newblock A survey on video diffusion models.
\newblock \emph{ACM Computing Surveys}, 2023.

\bibitem[Xue et~al.(2022)Xue, Hang, Zeng, Sun, Liu, Yang, Fu, and
  Guo]{xue2022advancing}
Hongwei Xue, Tiankai Hang, Yanhong Zeng, Yuchong Sun, Bei Liu, Huan Yang,
  Jianlong Fu, and Baining Guo.
\newblock Advancing high-resolution video-language representation with
  large-scale video transcriptions.
\newblock \emph{IEEE/CVF Conference on Computer Vision and Pattern
  Recognition}, 2022.

\bibitem[Yang et~al.(2024)Yang, Teng, Zheng, Ding, Huang, Xu, Yang, Hong,
  Zhang, Feng, et~al.]{yang2024cogvideox}
Zhuoyi Yang, Jiayan Teng, Wendi Zheng, Ming Ding, Shiyu Huang, Jiazheng Xu,
  Yuanming Yang, Wenyi Hong, Xiaohan Zhang, Guanyu Feng, et~al.
\newblock Cogvideox: Text-to-video diffusion models with an expert transformer.
\newblock \emph{arXiv preprint arXiv:2408.06072}, 2024.

\bibitem[Yokoo(2021)]{yokoo2021contrastive}
Shuhei Yokoo.
\newblock Contrastive learning with large memory bank and negative embedding
  subtraction for accurate copy detection.
\newblock \emph{arXiv preprint arXiv:2112.04323}, 2021.

\bibitem[Zhang et~al.(2023)Zhang, Wang, Zhang, Zhao, Yuan, Qin, Wang, Zhao, and
  Zhou]{zhang2023i2vgen}
Shiwei Zhang, Jiayu Wang, Yingya Zhang, Kang Zhao, Hangjie Yuan, Zhiwu Qin,
  Xiang Wang, Deli Zhao, and Jingren Zhou.
\newblock I2vgen-xl: High-quality image-to-video synthesis via cascaded
  diffusion models.
\newblock \emph{arXiv preprint arXiv:2311.04145}, 2023.

\bibitem[Zhang et~al.(2024)Zhang, Long, Pan, Qiu, Yao, Cao, and
  Mei]{zhang2024trip}
Zhongwei Zhang, Fuchen Long, Yingwei Pan, Zhaofan Qiu, Ting Yao, Yang Cao, and
  Tao Mei.
\newblock Trip: Temporal residual learning with image noise prior for
  image-to-video diffusion models.
\newblock \emph{IEEE/CVF Conference on Computer Vision and Pattern
  Recognition}, 2024.

\bibitem[Zheng et~al.(2024)Zheng, Peng, Yang, Shen, Li, Liu, Zhou, Li, and
  You]{opensora}
Zangwei Zheng, Xiangyu Peng, Tianji Yang, Chenhui Shen, Shenggui Li, Hongxin
  Liu, Yukun Zhou, Tianyi Li, and Yang You.
\newblock Open-sora: Democratizing efficient video production for all, 2024.

\bibitem[Zhong et~al.(2021)Zhong, Lee, Zhang, and Klein]{zhong2021adapting}
Ruiqi Zhong, Kristy Lee, Zheng Zhang, and Dan Klein.
\newblock Adapting language models for zero-shot learning by meta-tuning on
  dataset and prompt collections.
\newblock \emph{Empirical Methods in Natural Language Processing}, 2021.

\end{thebibliography}
